\documentclass[sn-mathphys]{sn-jnl}
\usepackage[utf8]{inputenc}
\usepackage{graphicx}
\usepackage{booktabs} 
\usepackage{bm}
\usepackage{amsmath}
\usepackage[normalem]{ulem}

\usepackage{graphicx,amssymb}
\usepackage{multirow}
\usepackage{threeparttable}
\usepackage{verbatim}
\usepackage{hyperref}

\usepackage{epsfig}
\newtheorem{definition}{Definition}
\newtheorem{proposition}{Proposition}

\newtheorem{remark}{Remark}
\usepackage{enumitem}

\usepackage{graphicx}
\usepackage{subfig}

\title[Doubly Stochastic Models]{Stochastic Gradient Descent with Random Label Noises: Doubly Stochastic Models and Inference Stabilizer}

\author[1]{\fnm{Haoyi} \sur{Xiong}}\email{xionghaoyi@baidu.com}
\author*[1]{\fnm{Xuhong} \sur{Li}}\email{lixuhong@baidu.com}
\author[1]{\fnm{Boyang} \sur{Yu}}\email{boyangjlu1030@163.com}
\author[2]{\fnm{Dongrui} \sur{Wu}}\email{drwu@hust.edu.cn}
\author[3]{\fnm{Zhanxing} \sur{Zhu}}\email{zhanxing.zhu@gmail.com}
\author[4]{\fnm{Dejing} \sur{Dou}}\email{dejingdou@gmail.com}

\affil[1]{\orgname{Baidu Inc.}, \city{Beijing}, \country{China}}
\affil[2]{\orgname{Huazhong University of Science and Technology}, \city{Wuhan}, \country{China}}
\affil[3]{\orgname{Peking University}, \city{Beijing}, \country{China}}
\affil[4]{\orgname{BCG Greater China}, \city{Beijing}, \country{China}}

\newcommand\lxh[2]{{}{#2}}

\begin{document}

\abstract{
    Random label noises (or observational noises) widely exist in practical machine learning settings. 
    While previous studies primarily focus on the affects of label noises to the performance of learning, our work intends to investigate the implicit regularization effects of the label noises, under mini-batch sampling settings of stochastic gradient descent (SGD), with assumptions that label noises are unbiased. Specifically, we analyze the learning dynamics of SGD over the quadratic loss with unbiased label noises, where we model the dynamics of SGD as a stochastic differentiable equation (SDE) with two diffusion terms (namely a \emph{Doubly Stochastic Model}). While the first diffusion term is caused by mini-batch sampling over the (label-noiseless) loss gradients as many other works on SGD~\cite{zhu2018anisotropic,wu2020noisy}, our model investigates the second noise term of SGD dynamics, which is caused by mini-batch sampling over the label noises, as an implicit regularizer. Our theoretical analysis finds such implicit regularizer would favor some convergence points that could stabilize model outputs against perturbation of parameters (namely \emph{inference stability}). Though similar phenomenon have been investigated in~\cite{blanc2020implicit}, our work doesn't assume SGD as an Ornstein-Uhlenbeck like process and achieve a more generalizable result with convergence of approximation proved. 
    
    To validate our analysis, we design two sets of empirical studies to analyze the implicit regularizer of SGD with unbiased random label noises for deep neural networks training and linear regression. Our first experiment studies \emph{the noisy self-distillation tricks} for deep learning, where student networks are trained using the outputs from well-trained teachers with additive unbiased random label noises. Our experiment shows that the implicit regularizer caused by the label noises tend to select models with improved inference stability. We also carry out experiments on SGD-based linear regression with unbiased label noises, where we plot the trajectories of parameters learned in every step and visualize the effects of implicit regularization. The results backup our theoretical findings.

}

\maketitle

\section{Introduction}
Stochastic Gradient Descent (SGD) has been widely used as an effective way to train deep neural networks with large datasets~\cite{bottou1991stochastic}. While the mini-batch sampling strategy was firstly proposed to lower the cost of computation per iteration, it has been considered to incorporate an implicit regularizer preventing the learning process from converging to the local minima with poor generalization performance~\cite{zhang2017understanding,zhu2018anisotropic,jastrzkebski2017three,hoffer2017,keskar2016large}. To interpret such implicit regularization, one can model SGD as gradient descent (GD) with gradient noises caused by mini-batch sampling~\cite{bottou2018optimization}. Studies have demonstrated the potentials of 
such implicit regularization or gradient noises to improve the generalization performance of learning from both theoretical~\cite{mandt2017stochastic,chaudhari2018stochastic,hu2019quasi,simsekli2019tail} and empirical aspects~\cite{zhu2018anisotropic,hoffer2017,keskar2016large}. In summary, gradient noises keep SGD away from converging to the sharp local minima that generalizes poorly~\cite{zhu2018anisotropic,hu2019quasi,simsekli2019tail} and would select a flat minima~\cite{hochreiter1997flat} as the outcome of learning.

\begin{table}[]
\begin{tabular}{@{}ll@{}}
\toprule
Symbols & Definitions and Equations \\ \midrule
$x_i$, and $y_i=f^*(x_i)$ & the $i^{th}$ data point and true label\hfill\eqref{eq:settings}\\
$\Tilde{y}_i$, and $\varepsilon_i$ & the $i^{th}$ noisy label and the label noise. \hfill\eqref{eq:settings}\\
$f(x,\theta)$ & the output of neural network with parameter $\theta$ and input $x$.\hfill\eqref{eq:dols}\\
$\widehat{\theta}$ &  the estimator of parameters of a neural network.      \hfill \eqref{eq:dols}   \\
$L^*_i$ & ${L}_i(\theta)=\frac{1}{2}(f(x_i,\theta)-{y}_i)^2$  the loss based on a noiseless sample.\hfill\eqref{eq:dols} \\
$\Tilde{L}_i$       &  $\Tilde{L}_i(\theta)=\frac{1}{2}(f(x_i,\theta)-\Tilde{y}_i)^2$  the loss based on a noisy sample. \hfill\eqref{eq:dols}    \\ \hline
\multicolumn{2}{c}{SGD without assumptions on label noises}\\\hline
$B_k$ & the mini-batch of samples drawn by the $k^{th}$ step of SGD \hfill\eqref{eq:sgd} \\ 
$b=\vert B_k\vert$ &  the constant batch size of $B_k$ \hfill\eqref{eq:sgd} \\ 
$\theta_k$ & the $k^{th}$ step of SGD .  \hfill\eqref{eq:sgd} \\
$V_k$       &   SGD noise caused by mini-batch sampling of loss gradients.\hfill\eqref{eq:k-step-gradient-noise}         \\
$\eta$      &  the learning rate of SGD. \hfill\eqref{eq:sgd}         \\
$\mathrm{\Theta}(t)$ &  the continuous-time dynamics of SGD.\hfill\eqref{eq:sgd-sde}          \\
$z_k$      &   the random vector of standard Gaussian.  \hfill\eqref{eq:ir}      \\
$W(t)$   & the Brownian motion over time. \hfill\eqref{eq:sgd-sde}        \\ 

$\bar\theta_k$ & the $k^{th}$ step of discrete-time approximation to $\mathrm{\Theta}(t)$.  \hfill\eqref{eq:discrete-approx-sgd-sde} \\ \hline
\multicolumn{2}{c}{SGD with Unbiased Label Noises (ULN)}\\\hline
$\theta^\mathrm{ULN}_k$ & the $k^{th}$ step of SGD with unbiased label noises .  \hfill\eqref{eq:dsm} \\
$\xi^*_k$       & SGD noises thru. mini-batch sampling of TRUE loss gradients. \hfill\eqref{eq:dsm}\\
$\xi^\mathrm{ULN}_k$       & SGD noise thru. mini-batch sampling of unbiased label noises. \hfill\eqref{eq:dsm}          \\
$\Sigma_N^\mathrm{SGD}$ & the covariance matrix of the TRUE loss gradients. \hfill\eqref{eq:covariance},\eqref{eq:doublystochcon},\eqref{eq:doublystochdis}\\
$\Sigma_N^\mathrm{ULN}$ & the covariance matrix based on unbiased label noises. \hfill\eqref{eq:covariance},\eqref{eq:doublystochcon},\eqref{eq:doublystochdis}\\
$\mathrm{\Theta}^\mathrm{ULN}(t)$ & the continuous-time doubly stochastic model. \hfill\eqref{eq:doublystochcon}\\
$\bar\theta^\mathrm{ULN}_k$ & the $k^{th}$ step of discrete-time doubly stochastic model.  \hfill\eqref{eq:doublystochdis} \\
$W_1(t)$, and $W_2(t)$       & two independent Brownian motions over time. \hfill\eqref{eq:doublystochcon}         \\ 
$z_k$, and $z'_k$      &   two independent random vectors of standard Gaussian.  \hfill\eqref{eq:ir},\eqref{eq:doublystochdis}.         \\
$\mathrm{\Theta}^\mathrm{LNL}(t)$ & the continuous-time dynamics under Label-NoiseLess settings. \hfill\eqref{eq:lnl-con}\\ 

\bottomrule
\end{tabular}
\caption{{Key Symbols and Definitions}}
\label{tab:symbols}
\end{table}

In this work, we aim at investigating the influence of random label noises to the implicit regularization under mini-batch sampling of SGD. To simplify our research, we assume the training dataset as a set of vectors $\mathcal{D}=\{x_1,x_2,x_3,\dots,x_N\}$.
The label $\Tilde{y}_i$ for every vector $x_i\in \mathcal{D}$ is the noisy response of the true neural network $f^*(x)$ such that
\begin{equation}\label{eq:settings}
    \Tilde{y}_i=y_i+\varepsilon_i,\ y_i=f^*(x_i),\ \mathrm{ and }\  
    \mathbb{E}[\varepsilon_i] =0,\ \mathrm{var}[\varepsilon_i]=\sigma^2\ ,
\end{equation}
where the label noise $\varepsilon_i$ is assumed to be an independent zero-mean \textbf{random variable}.~
In our work, the random label noises can be either (1) drawn from probability distributions before training steps (but \lxh{dynamized}{re-sampled} by mini-batch sampling of SGD) or (2) realized by the random variables per training iteration~\cite{han2018co}. Thus learning is to \lxh{approximate}{estimate $\widehat\theta$ in} $f(x,\widehat\theta)$ \lxh{beats}{for approximating} $f^*(x)$, such that
\begin{equation}
    \widehat\theta\gets\underset{\forall\theta\in\mathbb{R}^d}{\mathrm{arg min}}\ \left\{\frac{1}{N}\sum_{i=1}^N\Tilde{L}_i(\theta):=\frac{1}{2N}\sum_{i=1}^N\left(f(x_i,\theta)-\Tilde{y}_i\right)^2\right\}.\label{eq:dols}
\end{equation}
\lxh{}{Note that we denote ${L}^*_i(\theta)=\frac{1}{2}(f(x_i,\theta)-{y}_i)^2$ as the loss based on a noiseless sample in this work.} Inspired by~\cite{hochreiter1997flat,zhu2018anisotropic}, our work studies how unbiased label noises $\varepsilon_i$ ($1\leq i\leq N$) would affect the ``selection'' of $\widehat\theta$ from possible solutions, in the viewpoint of learning dynamics~\cite{saxe2014exact} of SGD under mini-batch sampling~\cite{li2017stochastic,wu2019noisy,hu2019quasi}. For symbols used in this paper, please refer to Table~\ref{tab:symbols}.

\subsection{Backgrounds: SGD Dynamics and Implicit Regularization}
To analyze the SGD algorithm solving the problem in Eq~\eqref{eq:dols}, we follow settings in~\cite{li2017stochastic} and consider SGD as an algorithm that, in the $k^{th}$ iteration with the estimate $\theta_k$, it randomly picks up a $b$-length subset of samples from the training dataset i.e., $B_k\subset \mathcal{D}$, and estimates the mini-batch stochastic gradient $\frac{1}{b}\sum_{\forall x_i\in B_k}\nabla \Tilde{L}_i(\theta_k)$, then updates the estimate for $\theta_{k+1}$ based on $\theta_k$, as follow
\begin{equation}\label{eq:sgd}
    \theta_{k+1}\gets\left(\theta_k-\frac{\eta}{\vert B_k\vert}\sum_{\forall x_i\in B_k}\nabla\Tilde{L}_i(\theta_k)\right)\ ,
\end{equation}
\lxh{}{where $\eta$ refers to the step-size of SGD. Furthermore, we can derive the mini-batch sampled loss gradients into the combination of the full batch loss gradient and the noise, such that}
\begin{equation}\label{eq:sgd-der}
\begin{aligned}
&\frac{\eta}{\vert B_k\vert}\sum_{\forall x_i\in B_k}\nabla\Tilde{L}_i(\theta_k)\\
=&\frac{\eta}{N}\sum_{\forall x_i\in \mathcal{D}}\nabla\Tilde{L}_i(\theta_k)-\left(\frac{\eta}{N}\sum_{\forall x_i\in \mathcal{D}}\nabla\Tilde{L}_i(\theta_k)-\frac{\eta}{\vert B_k\vert}\sum_{\forall x_i\in B_k}\nabla\Tilde{L}_i(\theta_k)\right)\\
=&\frac{\eta}{N}\sum_{\forall x_i\in \mathcal{D}}\nabla\Tilde{L}_i(\theta_k)-\sqrt{\eta}V_k(\theta_k)\ ,
\end{aligned}
\end{equation}
where $\eta$ refers to the step-size of SGD, and $V_k(\theta_k)$ refers to a stochastic gradient noise term caused by mini-batch sampling. The noise would converge to zero with increasing batch size, as follow
\begin{equation}\label{eq:k-step-gradient-noise}
    V_k(\theta_k)=\sqrt{\eta}\left(\frac{1}{N}\sum_{\forall x_i\in \mathcal{D}}\nabla\Tilde{L}_i(\theta_k)-\frac{1}{\vert B_k\vert}\sum_{\forall x_i\in B_k}\nabla\Tilde{L}_i(\theta_k)\right) \to \mathbf{0}_d,\ \text{ as }\ B\to N\ .
\end{equation}
With $\mathbf{d}t=\eta\to 0$ and the constant batch size $b=\vert B_k\vert$, SGD algorithm would diffuse to a continuous-time dynamics $\mathrm{\Theta}(t)$ with a stochastic differential equation (SDE), with weak convergence~\cite{li2017stochastic,hu2019diffusion}, as follow,
\begin{equation}\label{eq:sgd-sde}
    \mathbf{d}\mathrm{\Theta} = -\frac{1}{N}\sum_{i=1}^N\nabla\Tilde{L}_i(\mathrm{\Theta})\mathbf{d} t+
    \left(\frac{\eta}{b}\Tilde{\Sigma}_N^\mathrm{SGD}(\mathrm{\Theta})\right)^\frac{1}{2}\mathbf{d}W(t)\ 
\end{equation}
where $W(t)$ is a standard Brownian motion in $\mathbb{R}^d$, and 
\lxh{Let}{let} \lxh{}{us} define $\Sigma_N^\mathrm{SGD}(\mathrm{\Theta})$ as the sample covariance matrix of loss gradients $\nabla L_i(\mathrm{\Theta})$ for $1\leq i\leq N$. \lxh{}{Please note that, for detailed derivations to obtain above continuous-time approximation and the assumptions, please refer to~\cite{li2017stochastic}}. We follow~\cite{li2017stochastic} and do not make low-rank assumptions on $\Tilde{\Sigma}_N^\mathrm{SGD}(\mathrm{\Theta})$. Through Euler discretization~\cite{li2017stochastic,chaudhari2018stochastic}, one can approximate SGD as $\bar\theta_k$ such that
\begin{equation}\label{eq:discrete-approx-sgd-sde}
\begin{aligned}
    &\bar\theta_{k+1}\gets\bar\theta_k-\frac{\eta}{N}\sum_{\forall x_i\in \mathcal{D}}\nabla\Tilde{L}_i(\bar\theta_k)+\sqrt{\eta}\xi_k(\bar\theta_k),\ \text{and}\\
    &\xi_k(\bar\theta_k)=\left(\frac{\eta}{b}\Tilde{\Sigma}_N^\mathrm{SGD}(\bar\theta_k)\right)^\frac{1}{2}z_k,\  z_k\sim\mathcal{N}(0,\mathbf{I}_d)\ .
\end{aligned}
\end{equation}
The implicit regularizer of SGD is $\xi_k(\bar\theta_k)=\left(\frac{\eta}{b}\Tilde{\Sigma}_N^\mathrm{SGD}(\bar\theta_k)\right)^\frac{1}{2} z_k$ which is data-dependent and controlled by the learning rate $\eta$ and batch size $B$~\cite{smith2018don}.~\cite{mandt2017stochastic,chaudhari2018stochastic,hu2019quasi} discussed SGD for varational inference and enabled novel applications to samplers~\cite{zhang2019cyclical,xiong2019sphmc}. To understand the effect to generalization performance,~\cite{zhu2018anisotropic,smith2018don} studied the escaping behavior from the sharp local minima~\cite{keskar2016large} and convergence to the flat ones.~\cite{jia2020information} discover the way that SGD could find a flat local \lxh{minima}{minimum} from information-theoretical perspectives and propose a novel regularizer to improve the performance.  Finally,~\cite{gidel2019implicit} studied regularization effects to linear DNNs and our previous work~\cite{wu2019noisy} proposed new multiplicative noises to interpret SGD and obtain stronger theoretical properties.

\subsection{Our Contributions}
In this work, we assume the unbiased random label noises $\varepsilon_i$ ($1\leq i\leq N$) and the mini-batch sampler of SGD are independent. 
When the random label noises have been drawn from probability distributions prior to the training procedure, SGD re-samples the label noises and generates a new type of data-dependent noises, in addition to the stochastic gradient noises of label-noiseless losses, through re-sampling label-noisy data and averaging label-noisy loss gradients of random mini-batchs~\cite{fang2018online,li2018statistical}.

Our analysis shows that under mild conditions, with gradients of label-noisy losses, SGD might incorporate an additional data-dependent noise term, complementing with the stochastic gradient noises~\cite{li2017stochastic,wu2019noisy} of label-noiseless losses, through re-sampling the samples with label noises~\cite{li2018statistical} or dynamically adding noises to labels over iterations~\cite{han2018co}.
We consider such noises as an implicit regularization caused by unbiased label noises, and interpret the effects of such noises as a solution selector of learning procedure. More specifically, this work has made unique contributions as follow.

\subsubsection{Doubly Stochastic Models} We reviewed the preliminaries~\cite{li2017stochastic,ali2018continuous,hu2019quasi,wu2019noisy} and \lxh{extent}{extended} the analytical framework in~\cite{li2017stochastic} to interpret the effects of unbiased label noises as an additional implicit regularizer on top of the continuous-time dynamics of SGD. Through discretizing the continuous-time dynamics of label-noisy SGD, we write discrete-time approximation to the learning dynamics, denoted as $\theta^\mathrm{ULN}_k$ for $k=1,2,\dots$,  as 
\begin{equation}\label{eq:dsm}
    \theta^\mathrm{ULN}_{k+1}\gets\theta^\mathrm{ULN}_k-\frac{\eta}{N}\sum_{i=1}^N\nabla L_i^*(\theta^\mathrm{ULN}_k)+\sqrt{\eta}\xi_k^*(\theta^\mathrm{ULN}_k)+\sqrt{\eta}\xi_k^\mathrm{ULN}(\theta^\mathrm{ULN}_k),
\end{equation}
where $L_i^*(\theta)=(f(x_i,\theta)-f^*(x_i))^2$ refers to the \emph{label-noiseless loss function} with sample $x_i$ and the true (noiseless) label $y_i$, the noise term $\xi_k^*(\theta)$ refers to the stochastic gradient noise~\cite{li2017stochastic} of label-noiseless loss function $L_i^*(\theta)$, then we can obtain the new implicit regularizer caused by the unbiased label noises (ULN) for $\forall \theta\in\mathbb{R}^d$, which can be approximated as follow
    \begin{equation}
        \xi_k^{\mathrm{ULN}}(\theta)\approx\left(\frac{\eta\sigma^2}{bN}\sum_{i=1}^N\nabla_\theta f(x_i,\theta)\nabla_\theta f(x_i,\theta)^\top\right)^{\frac{1}{2}}z_k,\ \text{and}\ z_k\sim\mathcal{N}(\mathbf{0}_d,\mathbf{I}_d)\ ,
        \label{eq:ir}
    \end{equation}
where $z_k$ refers to a random noise vector drawn from the standard Gaussian distribution, $\theta_k$ refers to the parameters of network in the $k^{th}$ iteration, $(\cdot)^{1/2}$ refers to the Chelosky decomposition of the matrix, $\nabla_\theta f(x_i,\theta)=\partial f(x_i,\theta)/\partial \theta$ refers to the gradient of the neural network output for sample $x_i$ over the parameter $\theta_k$, and $B$ and $\eta$ are defined as \lxh{the learning rate and the batch size}{the batch size and the learning rate} of SGD respectively. Obviously, the strength of such implicit regularizer is controlled by $\sigma^2$, $B$ and $\eta$.

\paragraph{Summary of Results}Section~\ref{sec:models} formulates the algorithm of SGD with unbiased random label noises as a stochastic dynamics based on two noise terms (\textbf{Proposition}~\ref{prop:mean-var}), derives the Continuous-time and Discrete-time Doubly Stochastic Models from SGD algorithms (\textbf{Definitions}~\ref{def:con-dsm} and~\ref{def:dis-dsm}), and provides approximation error bounds (\textbf{Proposition}~\ref{prop:conv-approx}). Proofs of two propositions are provided in~\ref{proof:pro1} and~\ref{proof:pro2}

\subsubsection{Inference Stabilizer as Implicit Regularizer} The regularization effects of unbiased random label noises should be
\begin{equation}
    \mathbb{E}_{z_k}\left\|\xi_k^{\mathrm{ULN}}(\theta_k)\right\|_2^2\approx\frac{\eta\sigma^2}{bN}\sum_{i=1}^N \|\nabla_\theta f(x_i,{\theta}_k)\|_2^2=\frac{\eta\sigma^2}{bN}\sum_{i=1}^N \left\|\frac{\partial }{\partial\theta}f(x_i,\theta_k)\right\|_2^2\ ,
\end{equation}
where $\nabla_\theta f(x,{\theta})$ refers to the gradient of $f$ over $\theta$ and the effects \lxh{is}{are} controlled by the batch size $B$ and the variance of label noises $\sigma^2$. Similar results have been obtained by assuming the deep learning algorithms have been driven by an Ornstein-Uhlenbeck like process~\cite{blanc2020implicit}, while our work does not rely on such assumption but is all based on our proposed \emph{Doubly Stochastic Models}.


\paragraph{Summary of Results}Section~\ref{sec:regularization} analyzes the implicit regularization effects of unbiased random label noises for SGD, where we conclude the implicit regularizer as a controller of the neural network gradient norm $\frac{1}{N}\sum_{i=1}^N \|\nabla_\theta f(x_i,{\theta})\|_2^2$ in the dynamics (\textbf{Proposition}~\ref{prop:regularization}). We then offer \textbf{Remarks}~\ref{rem:infstable}, ~\ref{rem:escconv} and~\ref{rem:pert}  to characterize the behaviors of SGD with unbiased label noises: (1) SGD would escape the local minimums with higher gradient norms, due to the larger perturbation driven by the implicit regularizer, (2) the strength of implicit regularization effects is controlled by the learning rate $\eta$ and batch size $b$, and (3) it is possible to tune the performance of SGD through adding and controlling the unbiased label noises, as low neural network gradient norms usually correspond to flat loss landscapes.

To validate our three remarks, Section~\ref{sec:noisy-distill} presents the experiments based on \emph{self-distillation with unbiased label noises}~\cite{zhang2019your,kim2020self} for deep neural networks, where we show that, under the teacher-student training setting, a well-trained model could escape the local minimum and converge to a new point with lower neural network gradient norms and better generalization performance through learning from itself noisy outputs (i.e., logit outputs add unbiased label noises). Section~\ref{sec:noisy-linear} present experiments based on SGD-based linear regression with unbiased label noises to visualize implicit regularization effects and the connection between the effects and the learning rate/batch size. Our visualization results show SGD-based linear regression with unbiased label noises would converge to a distribution of Gaussian-alike centered at the solution of linear regression and the (co-)variance of the distribution is controlled by the covariance of data samples as well as the learning rate and batch size. Our experiment results backup our theory.


\section{Related Work}


\paragraph{SGD Implicit Regularization for Ordinary Least Square (OLS)} The most recent and relevant work in this area is~\cite{ali2018continuous,ali2020implicit}, where the same group of authors studied the implicit regularization of gradient descent and stochastic gradient descent for OLS. They investigated an implicit regularizer of $\ell_2$-norm alike on the parameter, which regularizes OLS as a Ridge estimator with decaying penalty. Prior to these efforts, F. Bach and his group have studied the convergence of gradient-based solutions for linear regression with OLS and regularized estimators under both noisy and noiseless settings in~\cite{dieuleveut2017harder,marteau2019beyond,berthier2020tight}.

\paragraph{Langevin Dynamics and Gradient Noises}
\lxh{}{With similar agendas,~\cite{mandt2017stochastic,chaudhari2018stochastic,hu2019quasi} studied limiting behaviors of SGD (or steady-state of dynamics) from the perspectives of Bayesian/variational inference. They also promoted novel applications to stochastic gradient MCMC samplers~\cite{zhang2019cyclical,xiong2019sphmc}. Through connecting $\Sigma_N^\mathrm{SGD}(\theta)$ to the loss Hessian $1/N\nabla^2 L_i(\theta)$ in near-convergence regions,~\cite{zhu2018anisotropic} studied the escaping behavior from the sharp local minima, while~\cite{keskar2016large} discussed this issue in large-batch training settings. Furthermore,~\cite{smith2018don} discussed how learning rates and batch sizes would affect the generalization performance and flatness of optimization results. Finally,~\cite{gidel2019implicit} studied the implicit regularization on linear neural networks and ~\cite{wu2019noisy} proposed a new multiplicative noise model to interpret the gradient noises with stronger theoretical properties.}

\paragraph{Self-Distillation and Noisy Students}  
Self-distillation~\cite{zhang2019your,xie2020self,xu2020knowledge,kim2020self} has been examined as an effective way to further improve the generalization performance of well-trained models. Such strategies enable knowledge distillation using the well-trained ones as teacher models and optionally adding noises (e.g., dropout, stochastic depth, and label smoothing or potentially the label noises) onto training procedure of student models.

\paragraph{Discussion on the Relevant Work} \lxh{Though tremendous pioneering studies have been done in this area}
{Compared to above works}, we still make contributions in above three categories. First of all, this work characterizes the implicit regularization effects of label noises to SGD dynamics. Compared to~\cite{ali2018continuous,ali2020implicit} working on linear regression, \lxh{our model interpreted general learning tasks}{our proposed doubly stochastic model could be used to explained the learning dynamics of SGD with label noises for nonlinear neural networks}. Even from linear regression perspectives~\cite{ali2018continuous,ali2020implicit,berthier2020tight}, we precisely measured the gaps between SGD dynamics with and without label noises and provide an new example with numerical simulation to visualize the implicit regularization effects.

Compared to~\cite{lopez2016unifying,kim2020self}, our analysis emphasized \lxh{}{the} role of the implicit regularizer caused by label noises for model selection, where models with high inferential stability would be selected.~\cite{li2020gradient} is the most relevant work to us, where authors studied the early stopping of gradient descent under label noises via neural tangent kernel (NTK)~\cite{jacot2018neural} approximation. Our work made the analyze for SGD without assumptions for approximation such as NTK. 

In addition to NTK assumption,~\cite{blanc2020implicit} assumes the deep learning algorithms are driven by an Ornstein-Uhlenbeck (OU) like process and obtains similar results as the inference stabilizer (the third result of our research), while our work makes contribution through proposing \emph{Doubly Stochastic Models} and reach the conclusion in a different way. We also provide yet the first empirical results and evidences, based on commonly-used DNN architectures and benchmark datasets, to visualize the effects of implicit regularizers caused by the unbiased label noises in real-world settings. 
Please note that an earlier manuscript~\cite{xiong2021implicit} from us has been put on OpenReview with discussion, where external reviewers demonstrated their concerns--part of results has been investigated in~\cite{blanc2020implicit} and we didn't provide the results in a strong form (e.g., theorems or proofs). Hereby, this work shifts the main contributions from implicit regularization of label noises to the \emph{doubly stochastic models} with approximation error bounds and proofs. The implicit regularization effects could be estimated via doubly stochastic models directly without the assumption of OU process. 
To best of our knowledge, this work is the first to understand the effects of unbiased label noises to SGD dynamics, by addressing technical issues including implicit regularization, OLS, self-distillation, model selection, and the stability inference results.

\section{Double Stochastic Models for SGD with Unbiased Random Label Noises}\label{sec:models} 
In this section, we present SGD with unbiased random label noises, derive the Continuous-time/Discrete-time Doubly Stochastic Models, and provide convergence of approximation between models.

\subsection{Modeling Unbiased Label Noises in SGD}
In our research, SGD with Unbiased Random Label Noises refers to an iterative algorithm that updates the estimate incrementally from initialization ${\theta}^\mathrm{ULN}_0$. 
With mini-batch sampling and unbiased random label noises, in the $k^{th}$ iteration, SGD algorithm updates the estimate $\theta^\mathrm{ULN}_k$ using the stochastic gradient $\Tilde{\mathrm{g}}_k(\theta^\mathrm{ULN}_k)$ through a gradient descent rule
, such that
\begin{equation}\label{eq:sgduln}
    \theta^\mathrm{ULN}_{k+1}\gets \theta^\mathrm{ULN}_k-\eta \Tilde{\mathrm{g}}_k(\theta^\mathrm{ULN}_k)\ ,
\end{equation}
Specifically, in the $k^{th}$ iteration, SGD randomly picks up a batch of sample $B_k\subseteq\mathcal{D}$ to estimate the stochastic gradient, as follow
\begin{equation}
\begin{aligned}
    \eta\Tilde{\mathrm{g}}_k(\theta^\mathrm{ULN}_k) & =\frac{\eta}{\vert B_k\vert}\sum_{x_i\in B_k}\nabla \Tilde{L}_i(\theta^\mathrm{ULN}_k)\\
    &=\frac{\eta}{\vert B_k\vert}\sum_{x_i\in B_k}\left(\left(f(x_i,\theta_k^\mathrm{ULN})-y_i\right)-\varepsilon_i\right)\cdot\nabla_\theta f(x_i,\theta_k^\mathrm{ULN})\\
    &= \frac{\eta}{N}\sum_{i=1}^N\nabla {L}^*_i(\theta^\mathrm{ULN}_k)+\sqrt{\eta}\xi_{k}^*(\theta^\mathrm{ULN}_k)+\sqrt{\eta}\xi_k^\mathrm{ULN}(\theta^\mathrm{ULN}_k),
\end{aligned}
\end{equation}
where $\nabla {L}^*_i(\theta)$ for $\forall\theta\in\mathbb{R}^d$ refers to the loss gradient based on the label-noiseless sample $(x_i,y_i)$ and $y_i=f^*(x_i)$, ~$\xi^*_{k}(\theta)$ refers to stochastic gradient noises~\cite{li2017stochastic} through mini-batch sampling over the gradients of label-noiseless samples, and $\xi_{k}^\mathrm{ULN}(\theta)$ is an additional noise term caused by the mini-batch sampling and the \underline{U}nbiased Random \underline{L}abel \underline{N}oises, such that
\begin{equation}\label{eq:key-terms}
\begin{aligned}
\nabla {L}^*_i(\theta)&=\frac{\partial}{\partial\theta}\frac{(f(x_i,\theta)-f^*(x_i))^2}{2}=\left(f(x_i,\theta)-f^*(x_i)\right)\cdot\nabla f(x_i,\theta)\ ,\\
\xi^*_{k}(\theta)&=\frac{\sqrt{\eta}}{\vert B_k\vert}\sum_{x_j\in B_k}\left(\nabla L^*_j(\theta) -\frac{1}{N}\sum_{i=1}^N\nabla L^*_i(\theta)\right),\\
\xi_{k}^\mathrm{ULN}(\theta)&= - \frac{\sqrt{\eta}}{\vert B_k\vert}\ \sum_{x_j\in B_k} \varepsilon_j\cdot\nabla_\theta f(x_j,\theta)\ .
\end{aligned}
\end{equation}
\begin{proposition}[Mean and Variance of the Two Noise Terms]\label{prop:mean-var} The mean and variance of the noise terms $\xi^*_{k}(\theta)$ and $\xi_{k}^\mathrm{ULN}(\theta)$ should be the vector-value functions as follow
\begin{equation}
\begin{aligned}
& \mathbb{E}_{B_k}[\xi^*_{k}(\theta)]  = \mathbf{0}_d, \ & \text{and}\ & \mathrm{Var}_{B_k}[\xi^*_{k}(\theta)]=\frac{\eta}{\vert B_k\vert}\Sigma_{N}^\mathrm{SGD}(\theta) \\
& {\mathbb{E}}_{B_k,\varepsilon_i}[\xi_{k}^\mathrm{ULN}(\theta)]  = \mathbf{0}_d, \ & \text{and}\ & \mathrm{Var}_{B_k\varepsilon_i}[\xi_{k}^\mathrm{ULN}(\theta)] =\frac{\eta}{\vert B_k\vert}\Sigma_{N}^\mathrm{ULN}(\theta)\ . \\
\end{aligned}
\end{equation}
The two matrix-value functions  $\Sigma_N^\mathrm{SGD}(\theta)$ and $\Sigma_N^\mathrm{ULN}(\theta)$ over $\theta\in\mathbb{R}^d$  characterize the variance of noise vectors. When we assume the label noises and mini-batch sampling are independent, there has
\begin{equation}\label{eq:covariance}
\begin{aligned}
    \Sigma_N^\mathrm{SGD}(\theta)&=\frac{1}{N}\sum_{j=1}^N\left(\nabla L^*_j(\theta) -\frac{1}{N}\sum_{i=1}^NL^*_i(\theta)\right)\left(\nabla L^*_j(\theta) -\frac{1}{N}\sum_{i=1}^NL^*_i(\theta)\right)^\top\\
    \Sigma_N^\mathrm{ULN}(\theta)&=\frac{\sigma^2}{N}\sum_{j=1}^N\nabla_\theta f(x_j,\theta)\nabla_\theta f(x_j,\theta)^\top\  \text{ as }\ \mathrm{var}[\varepsilon_j] =\sigma^2\ .
    \end{aligned}
\end{equation}
The two noise terms $\xi^*_{k}(\theta)$ and $\xi_{k}^\mathrm{ULN}(\theta)$ that are controlled by the learning rate \lxh{}{and} the batch size would largely influence the SGD dynamics. Please refer to~\ref{proof:pro1} for proofs.
\end{proposition}


With the mean and variance of two noise terms, we can easily formulate the learning dynamics of SGD with unbiased label noises as follows.


\subsection{Doubly Stochastic Models and Approximation}
We consider the SGD algorithm with unbiased random label noises in the form of gradient descent with additive data-dependent noise.
, such that $\theta_{k+1}=\theta_k-\frac{\eta}{N}\sum_{i=1}^N\nabla\Tilde{L}_i(\theta_k)+\sqrt{\eta}\Tilde{V}_k(\theta_k)$. 
When $\eta\to 0$, we assume the noise terms $\xi^*_{k}(\theta_k)$ and $\xi^\mathrm{ULN}_{k}(\theta_k)$ are independent, then we can follow the analysis in~\cite{hu2019diffusion} to derive the diffusion process of SGD with unbiased random label noises, denoted as $\mathrm{\Theta}^\mathrm{ULN}(t)$ over continuous-time $t\geq 0$. \lxh{Such that}{In this way,} we define the \emph{Doubly Stochastic Models} that characterizes the continuous-time dynamics of SGD with unbiased label noises as follows.

\begin{definition}[Continuous-Time Doubly Stochastic Models]\label{def:con-dsm} Given an SGD algorithm $\theta^\mathrm{ULN}_k$ defined and specified \lxh{}{in} Section 3.1, with $\eta=\mathbf{dt}$, we assume $B_k=B$ for $k=1, 2, 3\dots$ and formulate its continuous-time dynamics as
\begin{equation}\label{eq:doublystochcon}
\begin{aligned}
    \mathbf{d}{\mathrm{\Theta}}^\mathrm{ULN}= - \frac{1}{N}\sum_{i=1}^N \nabla L^*_i({\mathrm{\Theta}}^\mathrm{ULN})\mathbf{d}t &
    + \left(\frac{\eta}{b}\Sigma_N^\mathrm{SGD}({\mathrm{\Theta}}^\mathrm{ULN})\right)^\frac{1}{2}\mathbf{d}W_1(t) \\
    &  +\left(\frac{\eta}{b}\Sigma_N^\mathrm{ULN}({\mathrm{\Theta}}^\mathrm{ULN})\right)^\frac{1}{2}\mathbf{d}W_2(t)\ ,
\end{aligned}
\end{equation}
where ${W}_1(t)$ and ${W}_2(t)$ refer to two independent \lxh{Brownie}{Brownian} motions over time, $\mathbf{d}t={\eta}$ and $\mathrm{\Theta}^\mathrm{ULN}(0)=\theta^\mathrm{ULN}_0$. 
\end{definition}

Obviously, we can obtain the discrete-time approximation~\cite{li2017stochastic,chaudhari2018stochastic} to the SGD dynamics as \lxh{follow}{follows}.

\begin{definition}[Discrete-Time Doubly Stochastic Models]\label{def:dis-dsm}
We denote $\bar\theta^\mathrm{ULN}_k$ for $k=1,2,\dots$ as the discrete-time approximation to the \emph{Doubly Stochastic Models for SGD with Unbiased Label Noises}, which in the $k^{th}$ iteration behaves as
\begin{equation}\label{eq:doublystochdis}
\begin{aligned}
    \bar\theta^\mathrm{ULN}_{k+1}\gets \bar\theta^\mathrm{ULN}_k-\frac{\eta}{N}\sum_{i=1}^N\nabla L_i^*(\bar\theta^\mathrm{ULN}_k)&+\sqrt{\eta}\left(\frac{\eta}{b}\Sigma_N^\mathrm{SGD}({\bar\theta^\mathrm{ULN}_k})\right)^\frac{1}{2}z_k\\
    &+\sqrt{\eta}\left(\frac{\eta}{b}\Sigma_N^\mathrm{ULN}({\bar\theta^\mathrm{ULN}_k})\right)^\frac{1}{2}z'_k ,
\end{aligned}
\end{equation}
where $z_k$ and $z'_k$ are two independent $d$-dimensional random vectors drawn from a standard $d$-dimensional Gaussian distribution $\mathcal{N}(\mathbf{0}_d,\mathbf{I}_d)$ per iteration independently, and $\bar\theta^\mathrm{ULN}_0=\mathrm{\Theta}^\mathrm{ULN}(0)$. 
\end{definition}
The convergence between $\bar\theta^\mathrm{ULN}_k$ and $\mathrm{\Theta}^\mathrm{ULN}(t)$ is tight when $t=k\eta$ and the convergence bound is as follow.


\begin{proposition}[Convergence of Approximation]\label{prop:conv-approx} Let $T \ge 0$. Let $\Sigma_N^\mathrm{SGD}(\theta)$ and $\Sigma_N^\mathrm{ULN}(\theta)$ be the two diffusion matrices defined in Eq.~15. Assume that
\begin{enumerate}
    \item[A1] There exists some $M>0$ such that $\underset{_{i=1,2,...,N}}{\max}\{(\|\nabla L^*_i(\theta)\|_2)\}\leq M$ and $\underset{_{i=1,2,...,N}}{\max}\{(\|\nabla_\theta f(x_i;\theta)\|_2)\}\leq M$;
    \item[A2] There exists some $L>0$ such that $\nabla L_i(\theta)$ and $\nabla_\theta f(x,\theta)$ for $\forall x\in\mathcal{D}$ are Lipschitz continuous with bounded Lipschitz constant $L>0$ uniformly for all $i=1,2,...,N$.
\end{enumerate}
The continuous-time dynamics of SGD with unbiased label noises (ULN), denoted as $\mathrm{\Theta}^\mathrm{ULN}(t)$ in Eq.~\eqref{eq:doublystochcon}, is with order $1$ strong approximation to the discrete time SGD dynamics $\bar\theta_k^\mathrm{ULN}$ in Eq.~\eqref{eq:doublystochdis}. I.e., there exist a constant $C$ independent on $\eta$ but depending on $\sigma^2$, $L$ and $M$ such that
\begin{equation}
    \mathbb{E} \|\mathrm{\Theta}^\mathrm{ULN}({k\eta}) - \bar\theta^\mathrm{ULN}_k \|^2 \le C \eta^2, \quad \text{ for all } 0\leq k\leq \lfloor T/\eta \rfloor.
\end{equation}
Please refer to~\ref{proof:pro2} for proofs.
\end{proposition}

\begin{remark}
With above strong convergence bound for approximation, we can consider $\bar\theta^\mathrm{ULN}_k$ -- the solution of Eq.~\eqref{eq:doublystochdis} -- as a tight approximation to the SGD algorithm with unbiased label noises based on the same initialization. A tight approximation to the noise term $\xi_k^\mathrm{ULN}(\theta)$ (defined in Eq.~\eqref{eq:key-terms}) could be as follow 
\begin{equation}
  \xi_k^\mathrm{ULN}(\theta)\approx\left(\frac{\eta}{b}\Sigma_N^\mathrm{ULN}(\theta)\right)^\frac{1}{2}z'_k,\  \textbf{and}\ z'_k\sim\mathcal{N}(\mathbf{0_d},\mathbf{I_d})\ .
\end{equation}
We use such discrete-time iterations and approximations to the noise term $\xi_k^\mathrm{ULN}(\theta)$ to interpret the implicit regularization behaviors of the SGD with unbiased label noises algorithm $\theta^\mathrm{ULN}_k$ accordingly.
\end{remark}


\section{Implicit Regularization Effects to Neural Networks}\label{sec:regularization}
In this section, we use our model to interpret the regularization effects of \emph{SGD with unbiased label noises} for general neural networks, without assumptions on the structures of neural networks.

\subsection{Implicit Regularizer Influenced by Unbiased Random Label Noises} Compared the stochastic gradient with unbiased random label noises $\Tilde{g}_k(\theta)$ and the stochastic gradient based on the label-noiseless losses, we find an additional noise term $\xi_k^\mathrm{ULN}(\theta)$ as the \textbf{implicit regularizer}. 

To interpret $\xi_k^\mathrm{ULN}(\theta)$, we first define the diffusion process of SGD based on \underline{L}abel-\underline{N}oise\underline{L}ess losses i.e., $L_i^*(\theta)$ for $1\leq i\leq N$ as 
\begin{equation}\label{eq:lnl-con}
    \mathbf{d}\mathrm{\Theta}^\mathrm{LNL}=-\frac{1}{N}\sum_{i=1}^N\nabla L_i^*(\mathrm{\Theta}^\mathrm{LNL})\mathbf{d}t+\left(\frac{\eta}{b}\Sigma_N^\mathrm{SGD}(\mathrm{\Theta}^\mathrm{LNL})\right)^\frac{1}{2}\mathbf{W}(t)\ .
\end{equation}
Through comparing $\mathrm{\Theta}^\mathrm{ULN}(t)$ with $\mathrm{\Theta}^\mathrm{LNL}(t)$, the effects of $\xi_k^\mathrm{ULN}(\mathrm{\Theta})$ over contin-uous-time form should be $\sqrt{\eta/B}(\Sigma_N^\mathrm{ULN}(\mathrm{\Theta}))^{1/2}\mathbf{d}W(t)$. Then, in discrete-time, we could get results as follow.

\begin{proposition}[The implicit regularizer $\xi_k^\mathrm{ULN}(\theta)$]\label{prop:regularization}
The implicit regularizer of SGD with unbiased random label noises could be approximated as follow,
\begin{equation}
    \xi_k^{\mathrm{ULN}}(\theta)\approx\left(\frac{\sigma^2\eta}{bN}\sum_{i=1}^N\nabla_\theta f(x_i,\theta)\nabla_\theta f(x_i,\theta)^\top\right)^{\frac{1}{2}}z_k,\ \text{and}\ z_k\sim\mathcal{N}(\mathbf{0}_d,\mathbf{I}_d)\ .
\end{equation}
In this way, we can estimate the expected regularization effects of the implicit regularizer $\|\xi_k^{\mathrm{ULN}}(\theta)\|_2$ as follow,
\begin{equation}\label{eq:strength}
\begin{aligned}
\mathbb{E}_{z_k}\|\xi_k^{\mathrm{ULN}}(\theta)\|_2^2&=\frac{\eta\sigma^2}{bN}\sum_{i=1}^N\left\|\nabla_\theta f(x_i,\theta)\right\|_2^2 . 
\end{aligned}
\end{equation}
Please refer to~\ref{proof:pro3} for proofs.
\end{proposition}
We thus conclude that the effects of implicit regularization caused by unbiased random label noises for SGD is proportional to $\frac{1}{N}\sum_{i=1}^N\left\|\nabla_\theta f(x_i,\theta)\right\|_2^2 $ the average gradient norm of the neural network $f(x,\theta)$ over samples.



\subsection{Understanding the Unbiased Label Noises as an Inference Stabilizer} 
Here we extend the existing results on SGD~\cite{zhu2018anisotropic,wu2018sgd} to understand \textbf{Proposition 3} and obtain remarks as  follows.
\begin{remark}[Inference Stability]\label{rem:infstable} In the partial derivative form, the gradient norm could be written as 
    \begin{equation}\label{eq:infstable}
    \frac{1}{N}\sum_{i=1}^N\|\nabla_\theta f(x_i,\theta)\|_2^2=\frac{1}{N}\sum_{i=1}^N\|\frac{\partial}{\partial\theta}f(x_i,\theta)\|_2^2
    \end{equation} 
characterizes the variation of neural network output $f(x,\theta)$ based on samples $x_i$ (for $1\leq i\leq N$) over the parameter interpolation around the point $\theta$. Lower $\frac{1}{N}\sum_{i=1}^N\|\nabla_\theta f(x_i,\theta)\|_2^2$ \lxh{comes}{leads to} higher stability of neural network $f(x,\theta)$ outputs against the (random) perturbations over parameters. 
\end{remark}

\begin{remark}[Escape and Converge]\label{rem:escconv} When the noise $\xi_k^{\mathrm{ULN}}(\theta)$ is $\theta$-dependent (section 4 would present a special case that $\xi_k^{\mathrm{ULN}}(\theta)$ is $\theta$-independent with OLS), we follow~\cite{zhu2018anisotropic} and suggest that the implicit regularizer helps SGD escape from the point $\Tilde{\theta}$ with high neural network gradient norm $\frac{1}{N}\sum_{i=1}^N\|\nabla_\theta f(x_i,\Tilde{\theta})\|_2^2$, as the scale of noise $\xi_k^{\mathrm{ULN}}(\Tilde{\theta})$ is large. Reciprocally, we follow~\cite{wu2018sgd} and suggest that when the SGD with unbiased random label noises converges, \lxh{the converging point $\theta^*$ should be with a vanishing $\frac{1}{N}\sum_{i=1}^N\|\nabla_\theta f(x_i,\theta^*)\|_2^2$}{the algorithm would converge to a point $\theta^*$ with small $\frac{1}{N}\sum_{i=1}^N\|\nabla_\theta f(x_i,\theta^*)\|_2^2$}.  Similar results have been obtained in~\cite{blanc2020implicit} when assuming the deep learning algorithms are driven by an Ornstein–Uhlenbeck process.
\end{remark}

    
\begin{remark}[Performance Tuning]\label{rem:pert}
Considering $\eta\sigma^2/B$ as the coefficient balancing the implicit regularizer and vanilla SGD, one can regularize/penalize the SGD learning procedure with the fixed $\eta$ and $B$ more fiercely using a larger $\sigma^2$. More specifically, we could expect to obtain a regularized solution with lower $\frac{1}{N}\sum_{i=1}^N\|\nabla_\theta f(x_i,\theta)\|_2^2$ or higher inference stability of neural networks, as regularization effects become stronger when $\sigma^2$ increases.
\end{remark}

\section{Experiments on Self-Distillation with Unbiased Label Noises}\label{sec:noisy-distill}
The goal of this experiment is to understand \textbf{Proposition 3} and\emph{Remarks. 3 \& 4}, i.e., examining (1) whether the unbiased label noises would lower the gradient norm of the neural networks; (2) whether such unbiased label noises would improve the performance of neural networks; and (3) whether one can carry out performance tuning through controlling the variances of unbiased label noises, all in real-world deep learning settings. 

\subsection{Experiments Design}
To evaluate SGD with unbiased label noises, we design a set of novel experiments based on \emph{self-distillation with unbiased label noises}. In addition to learn from noisy labels directly, our experiment intends to train a (student) network from the noisy outputs of a (teacher) network in a quadratic regression loss, where the student network has been initialized from weights of the teacher one and unbiased label noises are given to the soft outputs of the teacher network randomly.

We aim to directly measure the gradient norm $\frac{1}{N}\sum_{i=1}^N\|\nabla_\theta f(x_i,\theta)\|_2^2$ of the neural network after every epoch to testify the SGD implicit regularization effects of unbiased label noises (i.e., \textbf{Proposition 3}). The performance comparisons among the teacher network, the student network (trained with unbiased label noises), and the student network (trained noiselessly) demonstrate the advantage of unbiased label noises in SGD for regression tasks (i.e., \emph{Remarks. 3 \& 4}).

Particularly, we design a set of novel experiments based on \emph{self-distillation with unbiased label noises} and elaborate in which way the proposed \emph{SGD with unbiased label noises} fits the settings of \emph{self-distillation with unbiased label noises}. Further, we introduce the goal of our empirical experiments with a list of expected evidences, then present the experiment settings for the empirical evaluation.  Finally, we present the experiment results with solid evidence to validate our proposals in this work. 


\subsection{Noisy Self-Distillation}
Given a well-trained model, Self-Distillation algorithms~\cite{zhang2019your,xu2020knowledge,kim2020self,xie2020self} intend to further improve the performance of a model through learning from the ``soft label'' outputs (i.e., logits) of the model (as the teacher). Furthermore, some practices found that the self-distillation could be further improved through incorporating certain randomness and stochasticity in the training procedure, namely noisy self-distillation, so as to obtain better generalization performance~\cite{xie2020self,kim2020self}. In this work, we study two well-known strategies for additive noises as follow.

\begin{enumerate}

\item \emph{Gaussian Noises.} Given a pre-trained model with $\mathbb{L}$-dimensional logit output, for every iteration of self-distillation, this simple first draws random vectors from a $\mathbb{L}$-dimensional Gaussian distribution $\mathcal{N}(\mathbf{0}_\mathbb{L},\sigma^2\mathbf{I}_\mathbb{L})$, then adds the vectors to the logit outputs of the model. It makes the student model learn from the noisy outputs. Note that in our analysis, we assume the output of the model is single dimension while, in self-distillation, the logit labels are with multiple dimensions. Thus, the diagonal matrix $\sigma^2\mathbf{I}_\mathbb{L}$ now refers to the complete form of the variances and $\sigma^2$ controls the scale of variances of noises. 

\item \emph{Symmetric Noises.}. Basically, this strategy is derived from ~\cite{han2018co} that generates noises through randomly swapping the values of logit output among the $\mathbb{L}$ dimensions. Specifically,     in every iteration of self-distillation, given a swap-probability $p$, every logit output (denoted as $y$ here) from the pre-trained model, and every dimension of logit output denoted as $y_l$, the strategy in probability $p$ swaps the logit value in the dimension that corresponds to $y_l$ with any other dimension $y_{m\neq l}$ in equal prior (i.e., in $(\mathbb{L}-1)^{-1}$ probability). In the rest $1-p$ probability, the strategy remains the original logit output there. In this way, the new noisy label $\Tilde{y}$ is with expectation $\mathbb{E}[\Tilde{y}]$ as follow,
    \begin{equation}
        \mathbb{E}[\Tilde{y}_l] = (1-p)\cdot y_l+\frac{p\cdot\sum_{\forall m\neq l} y_m}{\mathbb{L}-1} 
    \end{equation}
This strategy introduces explicit bias to the original logit outputs. However, when we consider the expectation $\mathbb{E}[\Tilde{y}]$ as the innovative soft label, then the random noise around the new soft label is still unbiased as $\mathbb{E}[\Tilde{y}-\mathbb{E}[\Tilde{y}]]=0$ for all dimensions.  Note that this noise is not the symmetric noises studied for robust learning~\cite{wang2019symmetric}.
\end{enumerate}
Thus, literately, our proposed \emph{SGD with Unbiased Label Noises} settings well fit the practice of noisy-self distillation.

\begin{figure}
	\centering
	\subfloat[Grad. Norm (SVHN)]{\includegraphics[width=0.33\textwidth]{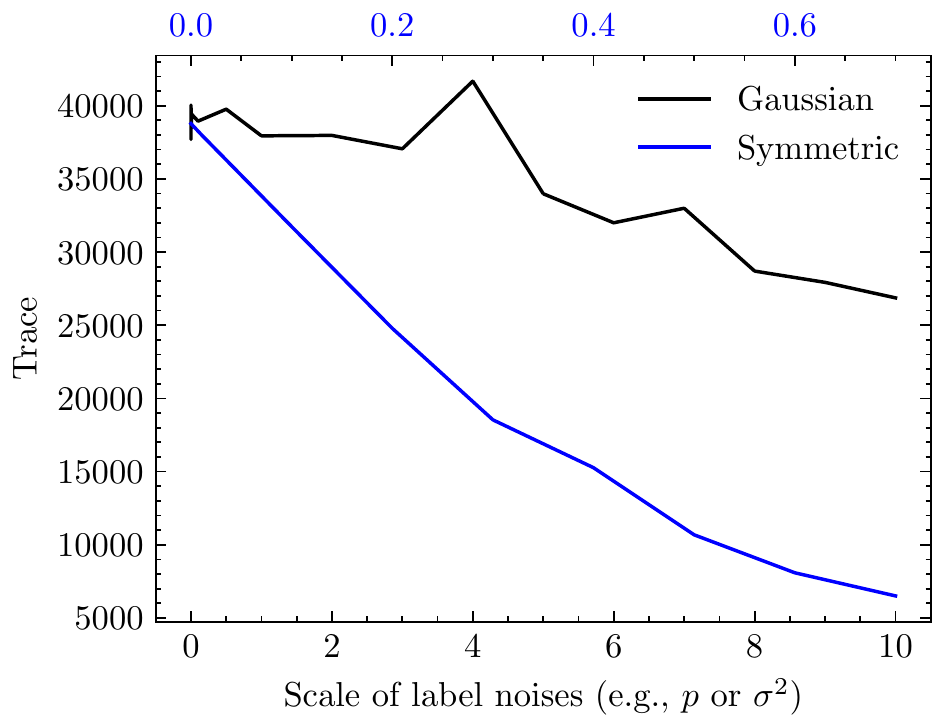}}
	\subfloat[Grad. Norm (CIFAR10)]{\includegraphics[width=0.33\textwidth]{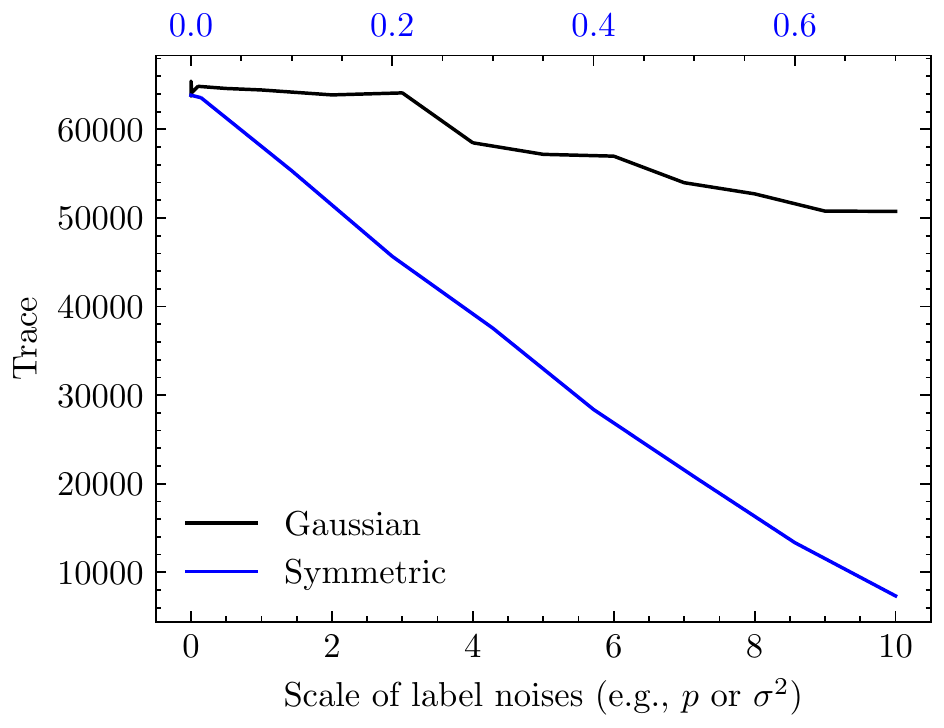}}
	\subfloat[Grad. Norm (CIFAR100)]{\includegraphics[width=0.33\textwidth]{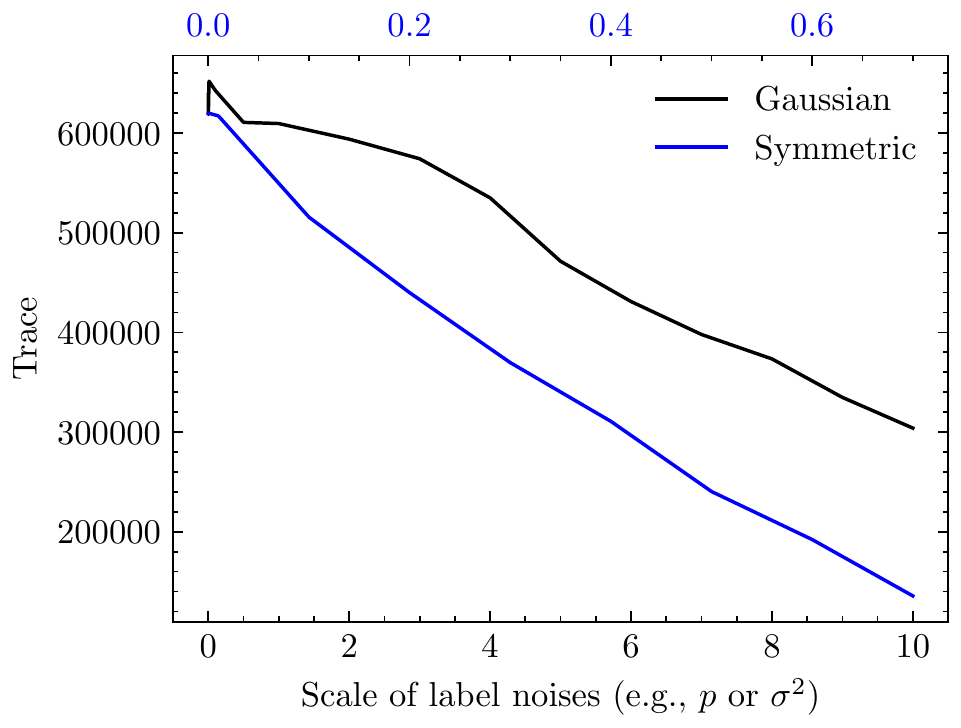}}\\ 
	\subfloat[Val. Acc. (SVHN)]{\includegraphics[width=0.33\textwidth]{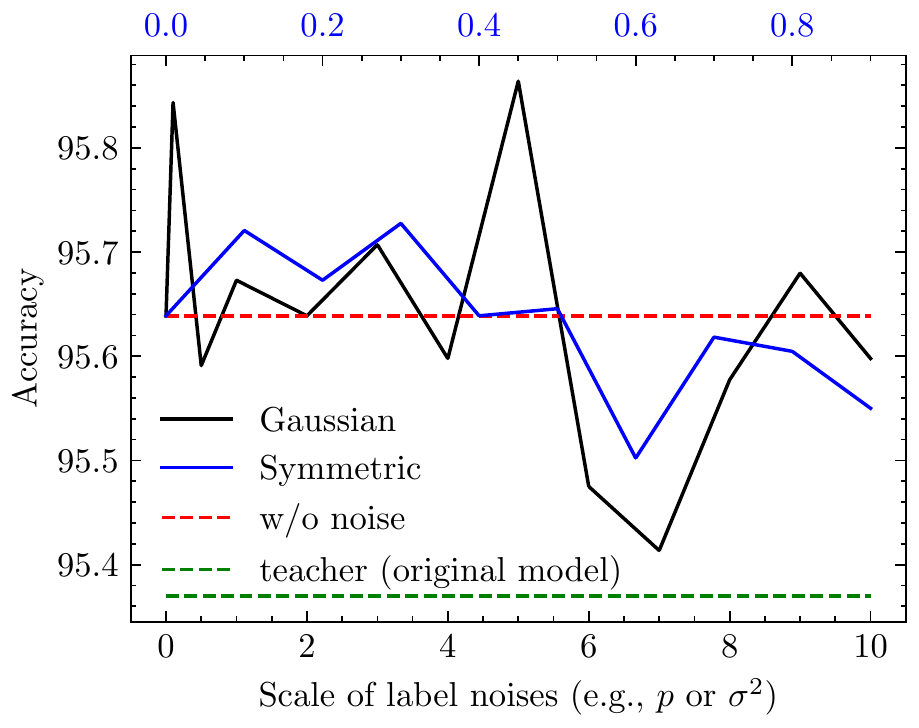}}
	\subfloat[Val. Acc. (CIFAR10)]{\includegraphics[width=0.33\textwidth]{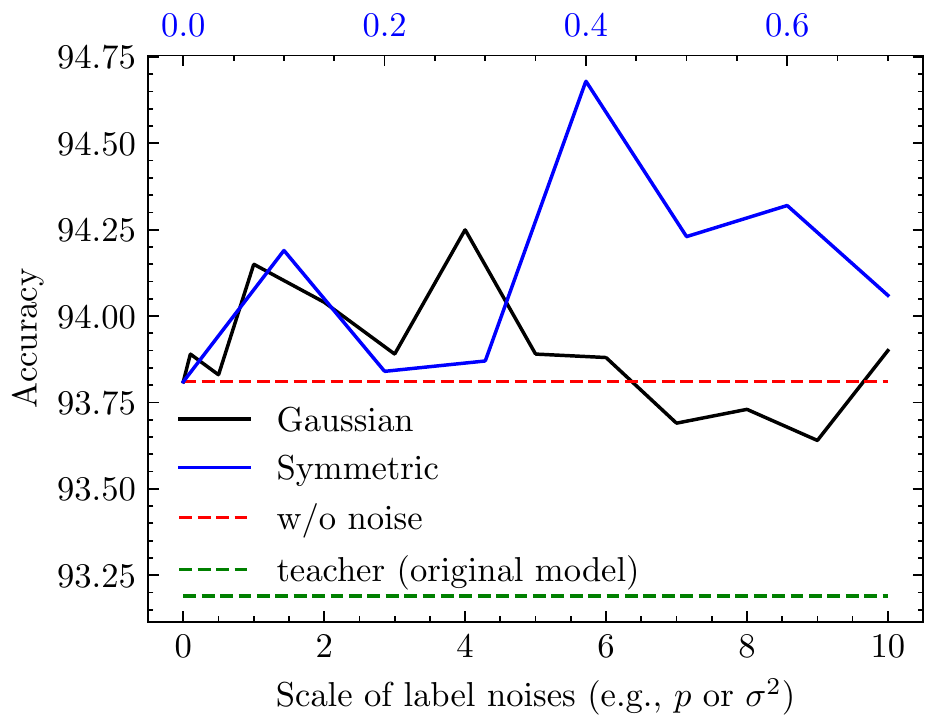}}
	\subfloat[Val. Acc. (CIFAR100)]{\includegraphics[width=0.33\textwidth]{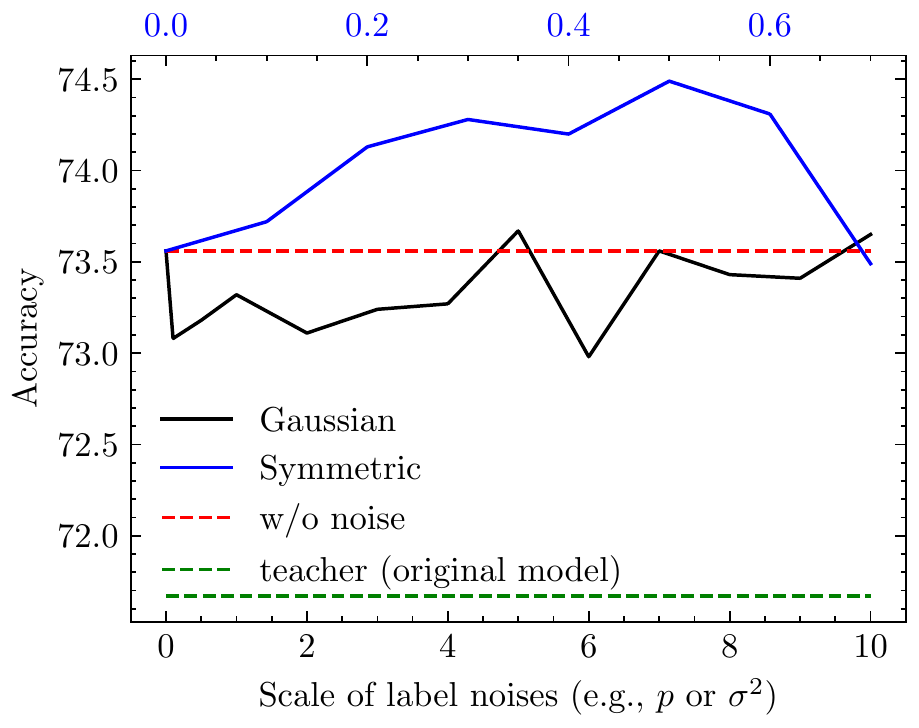}}\\ 
	\subfloat[Test Acc. (SVHN)]{\includegraphics[width=0.33\textwidth]{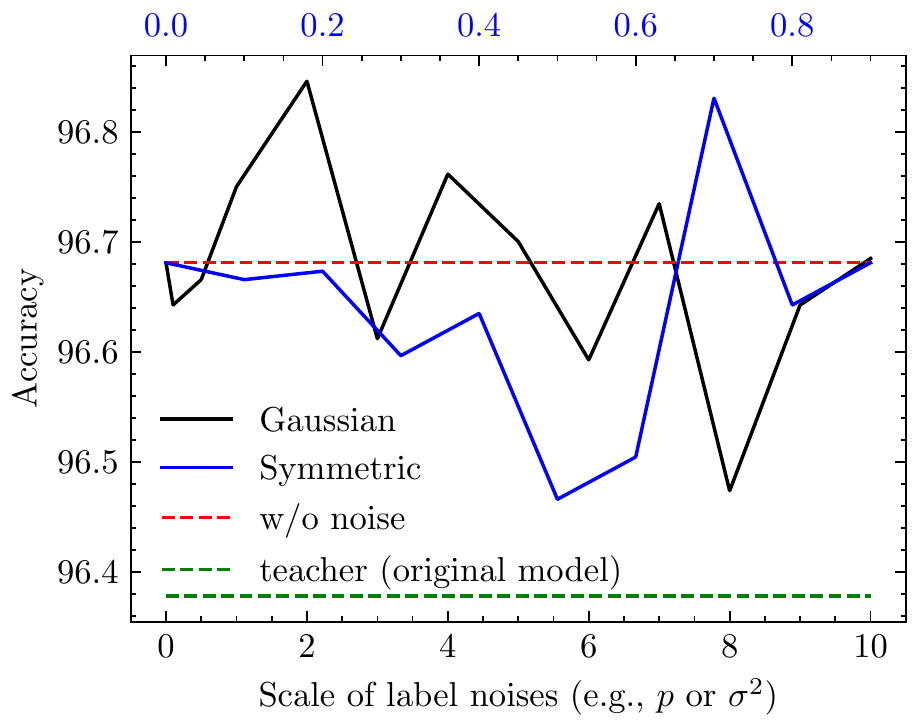}}
	\subfloat[Test Acc. (CIFAR10)]{\includegraphics[width=0.33\textwidth]{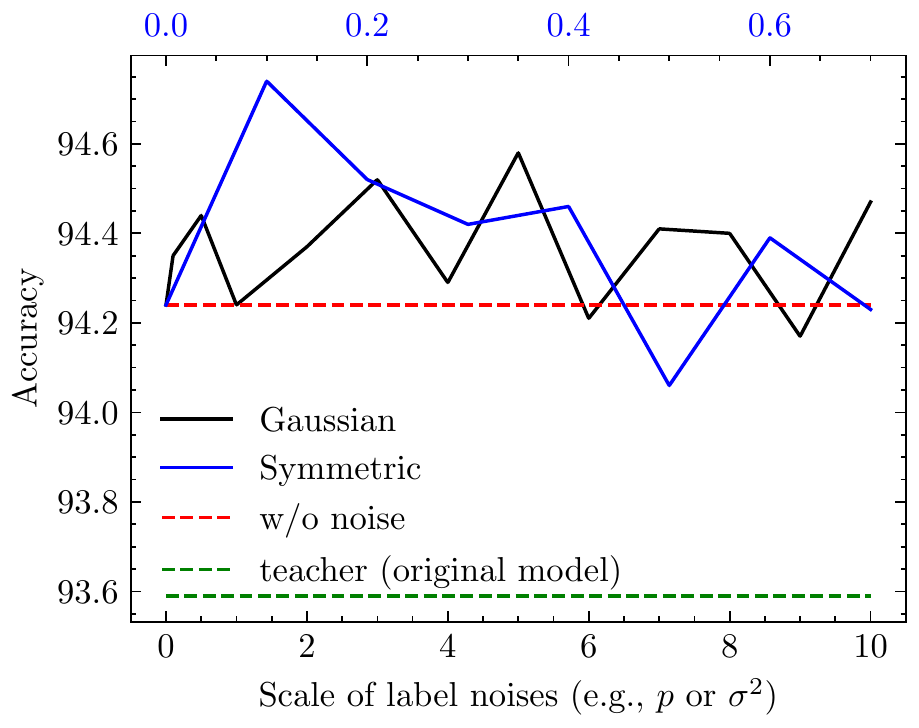}}
	\subfloat[Test Acc. (CIFAR100)]{\includegraphics[width=0.33\textwidth]{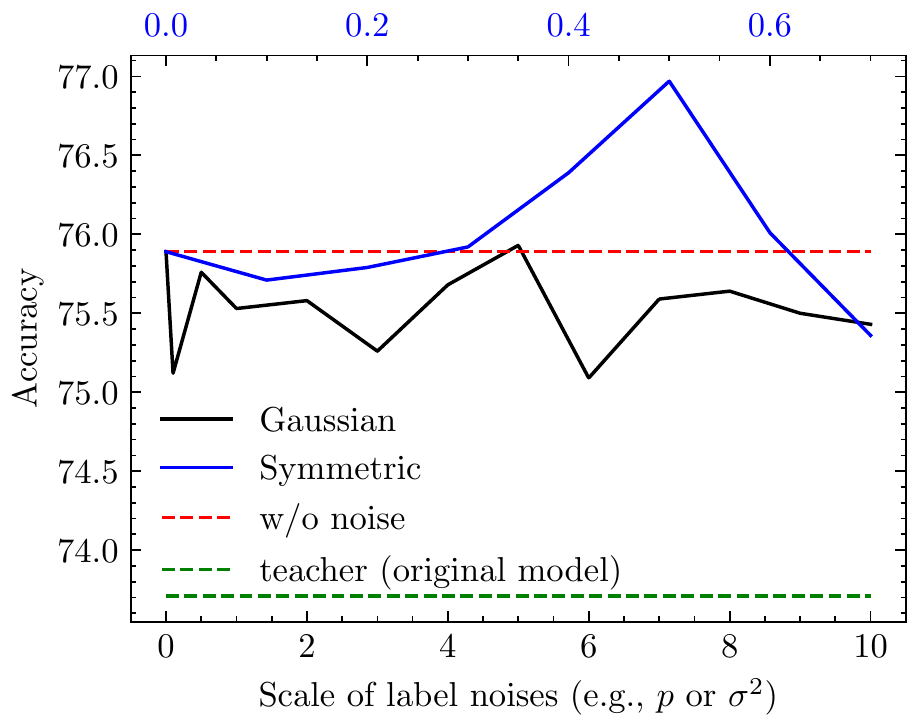}}
    \caption{Gradient Norms (corresponding to the inference stability), Validation Accuracy, and Testing Accuracy in Noisy Self-Distillation using ResNet-56 with varying scale of label noises (e.g., $p$ and $\sigma^2$) based on SVNH, CIFAR-10 and CIFAR-100 Datasets.}
	\label{fig:deep-models}
\end{figure}

\subsection{Datasets and DNN Models}
We choose the ResNet-56~\cite{he2016deep}, one of the most practical deep models, for conducting the experiments on three datasets: SVHN~\cite{netzer2011reading}, CIFAR-10 and CIFAR-100~\cite{krizhevsky2009learning}.
We follow the standard training procedure~\cite{he2016deep} for training a teacher model (original model).
Specifically we train the model from scratch for 200 epochs and adopt the SGD optimizer with batch size 64 and momentum 0.9.
The learning rate is set to 0.1 at the beginning of training and divided by 10 at 100$^{th}$ epoch and 150$^{th}$ epoch.
A standard weight decay with a small regularization parameter ($10^{-4}$) is applied.
As for noiseless self-distillation, we follow the standard procedure~\cite{Hinton2015distilling} for distilling knowledge from the teacher to a student of the same network structure.
The basic training setting is the same as training the teacher model.

For the settings of noisy self-distillation, we divide the original training set into a new training set (80\%) and a validation set (20\%). For clarity, we also present the results using varying scales of unbiased label noises on all three sets, where the original training set is used for training.

\begin{figure}
	\centering
	\subfloat[Train. Loss (SVHN)]{\includegraphics[width=0.33\textwidth]{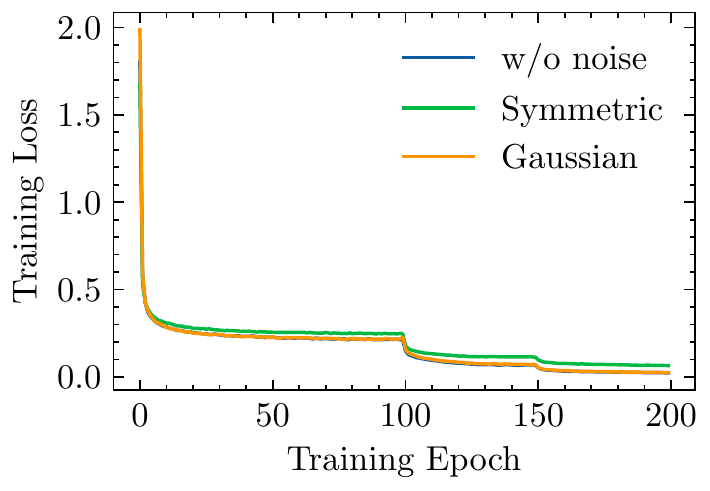}}
	\subfloat[Train. Loss (CIFAR-10)]{\includegraphics[width=0.33\textwidth]{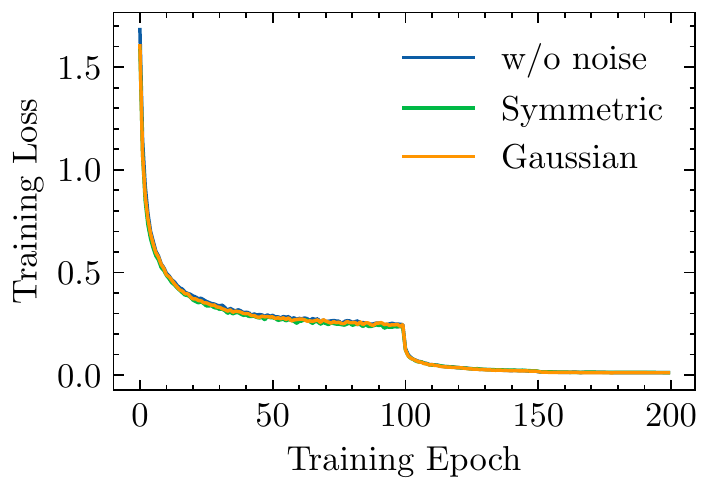}}
	\subfloat[Train. Loss (CIFAR-100)]{\includegraphics[width=0.33\textwidth]{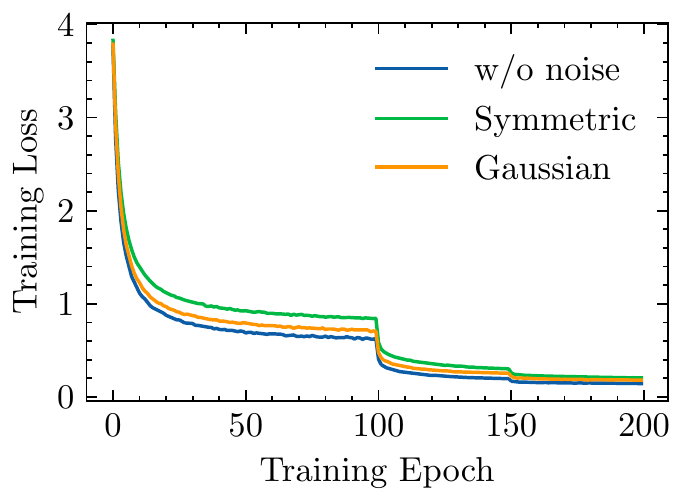}}
	
	\subfloat[Val. Loss (SVHN)]{\includegraphics[width=0.33\textwidth]{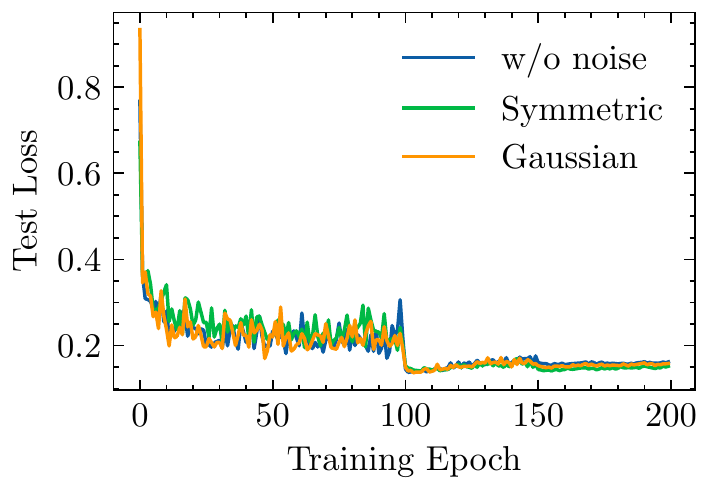}}
	\subfloat[Val. Loss (CIFAR-10)]{\includegraphics[width=0.33\textwidth]{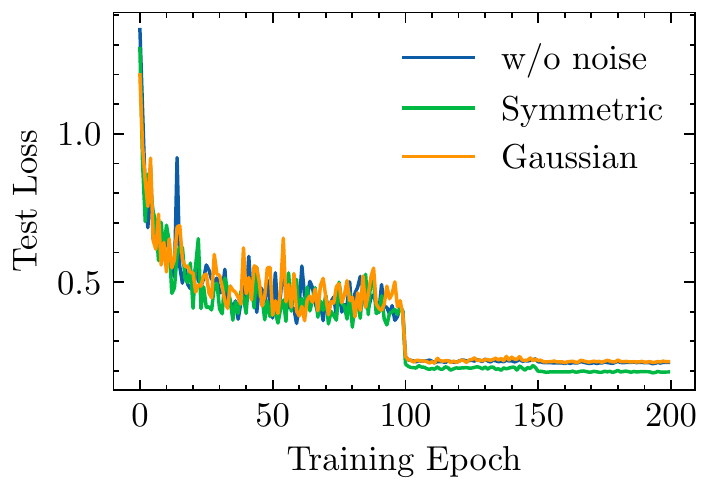}}
	\subfloat[Val. Loss (CIFAR-100)]{\includegraphics[width=0.33\textwidth]{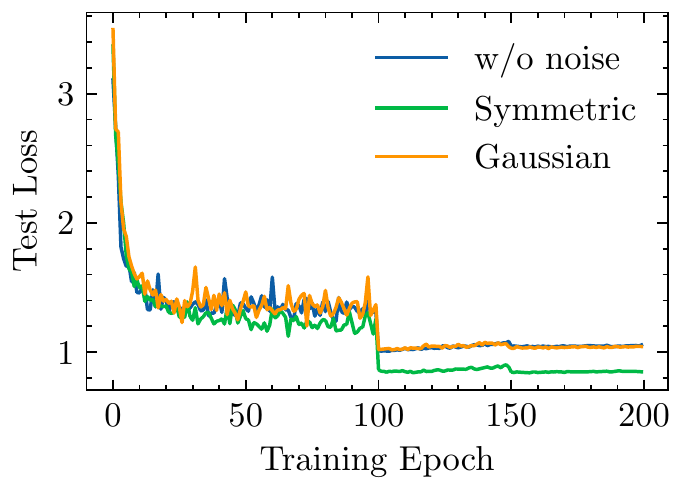}}
    \caption{Training and Validation Loss per Epoch during the Training Procedure}
	\label{fig:losses}
\end{figure}

\subsection{Experiment Results}
Figure~\ref{fig:deep-models} presents the results of above two methods with increasing scales of noises, i.e., increasing $\sigma^2$ for Gaussian noises and increasing $p$ for Symmetric noises. In Figure~\ref{fig:deep-models}(a)--(c), we demonstrate that the gradient norms of neural networks $\frac{1}{N} \sum_{i=1}^N\|\nabla_\theta f(x_i,\theta)\|_2^2$ decrease with growing $\sigma^2$ and $p$ for two strategies. The results backup our theoretical investigation 
, which means the model would be awarded high inferential stability, as the variation of neural network outputs against the potential random perturbation in parameters has been reduced by the regularization. In Figure~\ref{fig:deep-models}(d)--(f) and (g)--(i), we plot the validation and testing accuracy of the models obtained under noisy self-distillation. The results show that (1) student models are with lower gradient norms of neural networks $\frac{1}{N} \sum_{i=1}^N\|\nabla_\theta f(x_i,\theta)\|_2^2$ than teacher models, the gradient norm further decreases with increasing scale of noises (i.e., $\sigma^2$ and $P$); (2) some of models have been improved through noisy self-distillation compared to the teacher model, while noisy self-distillation could obtain better performance than noiseless self-distillation; and (3) it is possible to select noisily self-distilled models using validation accuracy for better overall generalization performance (in testing dataset). All results here are based on 200 epochs of noisy self-distillation.

We show the evolution of training and test losses during the entire training procedure, and compare the settings of adding no label noises, symmetric and Gaussian noises for noisy self-distillation. 
Figure \ref{fig:losses} presents the results on the three datasets, i.e., SVHN, CIFAR-10 and CIFAR-100 with the optimal scales of label noises on validation sets. It shows all algorithms would finally converge to a local minima with a training loss near to zero, while the local minimas searched by the SGD with Symmetric noise would be flatter with better generalization performance (especially for CIFAR-100 dataset).


\section{Experiments on Linear Regression with Unbiased Label Noises}\label{sec:noisy-linear} 
To validate our findings in linear regression settings, we carry out numerical evaluation using synthesize data to simply visualize the dynamics over iteration of SGD algorithms with label-noisy OLS and label-noiseless OLS. 

\subsection{Linear Regression with Unbiased Label Noises}
We here hope to see how unbiased label noises would influence SGD iterates for ordinary linear regression (OLS), where a simple quadratic loss function is considered for OLS, such that
\begin{equation}
   \widehat\beta_\mathrm{OLS} \gets \underset{\beta\in\mathbb{R}^d}{\mathrm{arg min}}\ \left\{ \frac{1}{N}\sum_{i=1}^N\Tilde{L}_i(\beta):=\frac{1}{2N}\sum_{i=1}^N\left(x_i^\top\beta-\Tilde{y}_i\right)^2\right\}\ ,\label{eq:linear-net}
\end{equation}
where samples are generated through $\Tilde{y}_i=x_i^\top\beta^*+\varepsilon_i$, $\mathbb{E}[\varepsilon_i]=0$ and $\mathrm{var}[\varepsilon_i]=\sigma^2$. Note that in this section, we replace the notation of $\theta$ with $\beta$ to present the parameters of linear regression models.

Let \lxh{}{us} combine Eq.~\eqref{eq:linear-net} and Eq.~\eqref{eq:sgduln}. We write the SGD for Ordinary Least Squares with Unbiased Label Noises as the iterations $\beta_k^\mathrm{ULN}$ for $k=1,2,3\dots$ as follow
    \begin{equation}
    \beta^\mathrm{ULN}_{k+1}\gets\beta^\mathrm{ULN}_k-\frac{\eta}{N}\sum_{i=1}^N\nabla L_i^*(\beta^\mathrm{ULN}_k)+\sqrt{\eta}\xi_k^*(\beta^\mathrm{ULN}_k)+\sqrt{\eta}\xi_k^\mathrm{ULN}(\beta^\mathrm{ULN}_k),
\end{equation}
where $\nabla {L}^*_i(\beta)$ for $\forall\theta\in\mathbb{R}^d$ refers to the loss gradient based on the label-noiseless sample $(x_i,y_i)$ and $y_i=x_i^\top\beta^*$, ~$\xi^*_{k}(\beta)$ refers to stochastic gradient noises caused by mini-batch sampling over the gradients of label-noiseless samples, and $\xi_{k}^\mathrm{ULN}(\beta)$ is an additional noise term caused by the mini-batch sampling and the unbiased label noises, such that
\begin{equation}\label{eq:key-terms}
\begin{aligned}
\nabla {L}^*_i(\beta)&=\frac{\partial}{\partial\beta}\frac{(x_i^\top\beta-x_i^\top\beta^*)^2}{2}=x_ix_i^\top(\beta-\beta^*) ,\\
\xi^*_{k}(\beta)&=\frac{\sqrt{\eta}}{\vert B_k\vert}\sum_{x_j\in B_k}\left(\nabla L^*_j(\beta) -\frac{1}{N}\sum_{i=1}^NL^*_i(\beta)\right),\\
\xi_{k}^\mathrm{ULN}(\beta)&= - \frac{\sqrt{\eta}}{\vert B_k\vert}\ \sum_{x_j\in B_k} \varepsilon_j\cdot x_j\ .
\end{aligned}
\end{equation}
We denote the \emph{sample covariance matrix} of $N$ samples as 
\begin{equation}
    \bar\Sigma_N=\frac{1}{N}\sum_{i=1}^{N}x_ix_i^\top\ .
\end{equation}
According to \emph{Remark 1} for implicit regularization in the general form, we can write the implicit regularizer of SGD with the random label noises for OLS as,
\begin{equation}
\begin{aligned}
\xi_k^\mathrm{ULN}(\beta)\approx\left(\frac{\eta\sigma^2}{B}\bar\Sigma_N\right)^\frac{1}{2}z'_k=\left({\frac{\eta\sigma^2}{bN}\sum_{i=1}^Nx_ix_i^\top}\right)^\frac{1}{2}
z'_k,\ \text{and}\ z'_k\sim\mathcal{N}(\mathbf{0}_d,\mathbf{I}_d),\ 
\end{aligned}\label{eq:oup}
\end{equation}
which is unbiased $\mathbb{E}[\xi_k^\mathrm{ULN}(\beta)]=\mathbf{0_d}$ with invariant covariance structure, and is independent with $\beta$ (the location) and $k$ (the time).

Let us combine \textbf{Proposition 2} and linear regression settings, we obtain the continuous-time dynamics for linear regression with unbiased label noises, denoted as $\beta^\mathrm{ULN}(t)$. According to~\cite{berthier2020tight,latz2021analysis}, we can see SGD and its continuous-time dynamics for noiseless linear regression (denoted as $\beta^\mathrm{LNL}_k$ and $\beta^\mathrm{LNL}(t)$) would asymptotically converge to the optimal solution $\beta^*$. As the additional noise term $\xi^\mathrm{ULN}$ is unbiased with an invariant covariance structure, when $t\to\infty$, we can simply conclude that  $\underset{t\to\infty}{\lim}\mathbb{E}\ \beta^\mathrm{ULN}(t)=\underset{t\to\infty}{\lim}\mathbb{E}\ \beta^\mathrm{LNL}(t)=\beta^*$, $\underset{t\to\infty}{\lim}\mathbf{d}\beta^\mathrm{ULN}(t)=(\frac{\eta\sigma^2}{B}\bar\Sigma_N)^{1/2}\mathbf{d}W(t)$. By definition of a distribution from a stochastic process, we could conclude  $\beta^\mathrm{ULN}(t)$ converges to a stationary distribution, such that
$\beta^\mathrm{ULN}(t)\sim\mathcal{N}(\beta^*,\frac{\eta\sigma^2}{B}\bar\Sigma_N)$, as $t\to\infty$.
.

\begin{remark}  Thus, with $k\to\infty$, the SGD algorithm for OLS with unbiased label noises would converge to a distribution of Gaussian-alike as follow
\begin{equation}
  \lim_{k\to\infty}\mathbb{E}\ [\beta_k^\mathrm{ULN}]=\beta^*, \text{and}\ \lim_{k\to\infty}\mathrm{Var}\ [\beta_k^\mathrm{ULN}] = \frac{\eta\sigma^2}{B}\bar\Sigma_N,
\end{equation}
The span and shape of the distribution are controlled by $\sigma^2$ and $\bar\Sigma_N$ when $\eta$ and $B$ are constant. 
\end{remark}

In this experiment, we hope to evaluate above remark using numerical simulations, so as to testify (1) whether the trajectories of $\beta_k^\mathrm{ULN}$ converges to a distribution of $\mathcal{N}(\beta^*,\frac{\eta\sigma^2}{B}\bar\Sigma_N)$; (2) whether the shape of convergence area could be controlled by the sample covariance matrix the data $\bar\Sigma_N$ of; and (3) whether the size of convergence area could be controlled by the variance of label noises $\sigma^2$.

\subsection{Experiment Setups}
In our experiments, we use 100 random samples realized from a 2-dimension Gaussian distribution $X_i\sim\mathcal{N}(\mathbf{0},\Sigma_{1,2})$ for $1\leq i\leq 100$, where $\Sigma_{1,2}$ is an symmetric covariance matrix controlling the random sample generation. To add the noises to the labels, we first \lxh{drawn}{draw} 100 copies of random noises from the normal distribution with the given variance $\varepsilon_i\sim\mathcal{N}(0,{\sigma}^2)$, then we setup the OLS problem with $(X_i,\Tilde{y}_i)$ pairs using $\Tilde{y}_i={X}_i^\top\beta^*+\varepsilon_i$ and $\beta^*=[1,1]^\top$ and various settings of ${\sigma}^2$ and $\Sigma_{1,2}$. We setup the SGD algorithms with the fixed learning rate $\eta=0.01$, and \lxh{bath}{batch} size $B=5$, with the total number of iterations $K=1,000,000$ to visualize the complete paths. 

\begin{figure}
    \subfloat[${\sigma}^2=0.25$]{\includegraphics[width=0.24\textwidth]{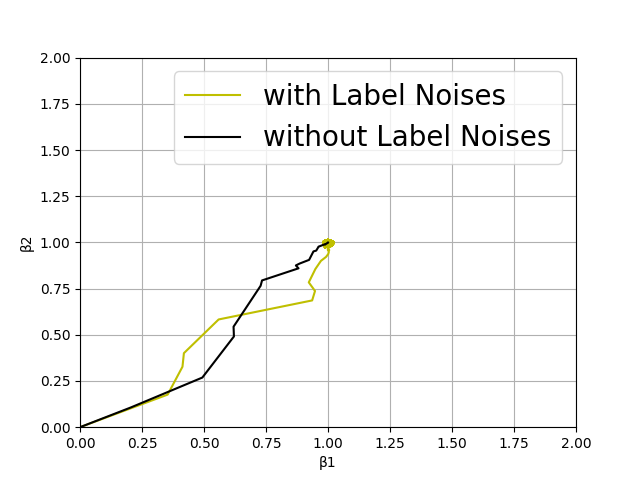}}\hfill
    \subfloat[${\sigma}^2=0.5$]{\includegraphics[width=0.24\textwidth]{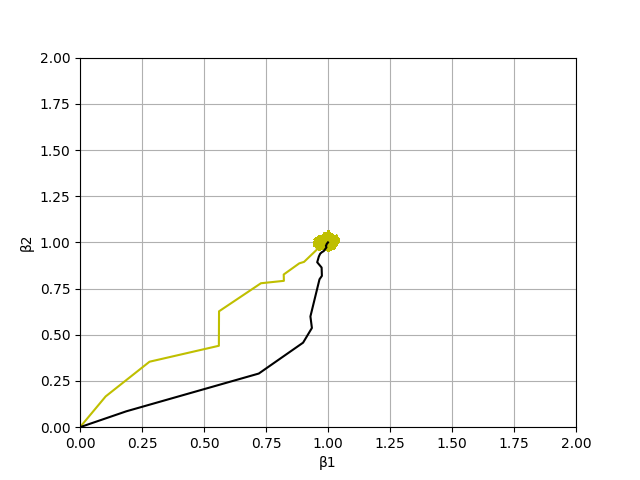}}\hfill
    \subfloat[${\sigma}^2=1.0$]{\includegraphics[width=0.24\textwidth]{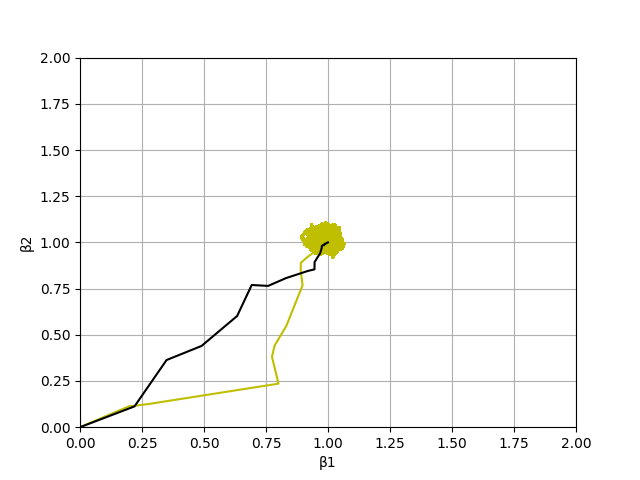}}\hfill
    \subfloat[${\sigma}^2=2.0$]{\includegraphics[width=0.24\textwidth]{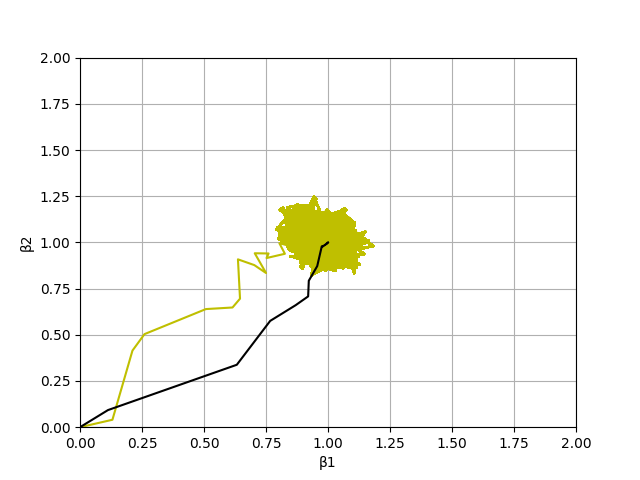}}\hfill
    \subfloat[$\Sigma_{1,2}=10\cdot\mathbf{I}_d$]{\includegraphics[width=0.24\textwidth]{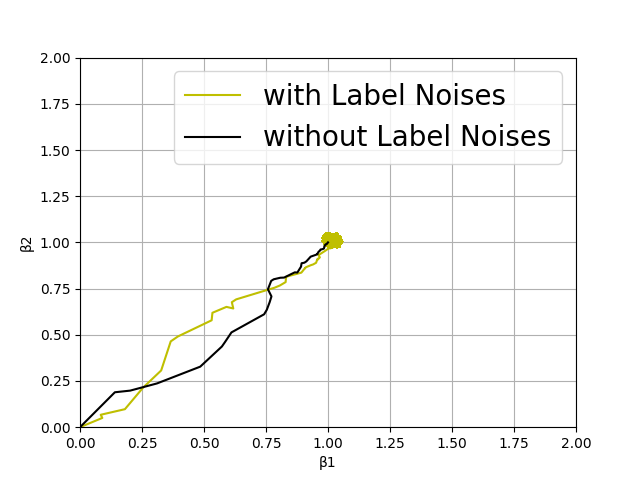}}\hfill
    \subfloat[$\Sigma_{1,2}=100\cdot\mathbf{I}_d$]{\includegraphics[width=0.24\textwidth]{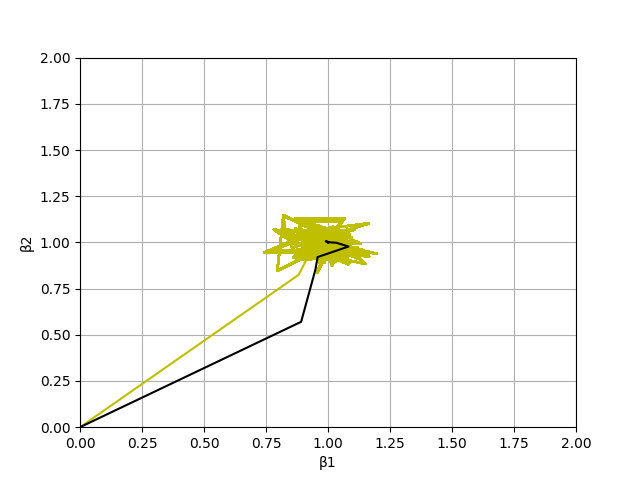}}\hfill
    \subfloat[$\Sigma_{1,2}=\Sigma^\mathrm{Ver}$]{\includegraphics[width=0.24\textwidth]{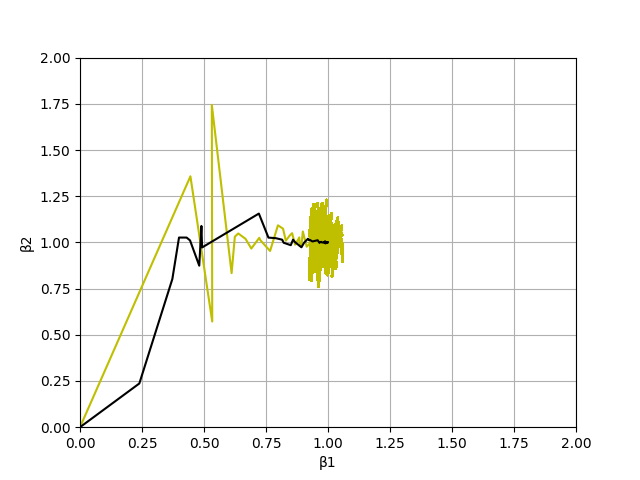}}\hfill
    \subfloat[$\Sigma_{1,2}=\Sigma^\mathrm{Hor}$]{\includegraphics[width=0.24\textwidth]{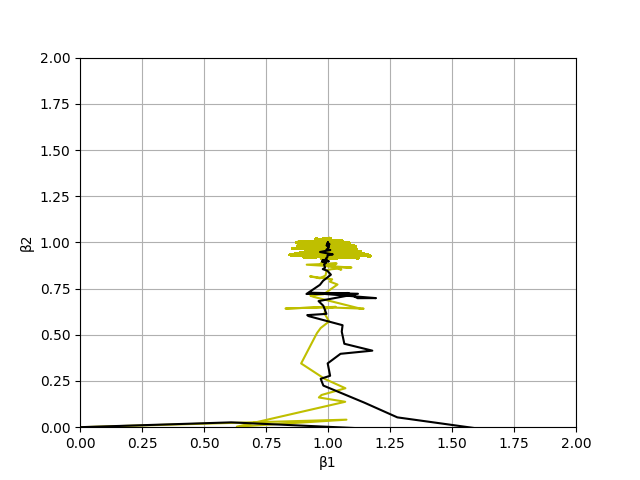}}\hfill
    \caption{Trajectories of SGD over OLS with and without Unbiased Random Label Noises using various $\Tilde{\sigma}^2$ and $\Sigma_{1,2}$ settings for (noisy) random data generation. For Figures~(a)--(d), the experiments are setup with a fixed $\Sigma_{1,2}=[[20,0]^\top,[0,20]^\top]$ and varying $\Tilde{\sigma}^2$. For Figures~(e)--(h), the experiments are setup with a fixed $\Tilde{\sigma}^2=0.5$ and varying $\Sigma_{1,2}$, where we set $\Sigma^\mathrm{Ver}=[[10,0]^\top,[0,100]^\top]$ and $\Sigma^\mathrm{Hor}=[[100,0]^\top,[0,10]^\top]$ to shape the converging distributions.}
    \label{fig:SGD-OLS}
\end{figure}

\subsection{Experiment Results}
Figure~\ref{fig:SGD-OLS} presents the results of numerical validations. In Figure~\ref{fig:SGD-OLS}(a)--(d), we gradually \lxh{increases}{increase} the variances of label noises $\sigma^2$ from $0.25$ to $2.0$, where we can observe (1) SGD over label-noiseless OLS converges to the optimal solution $\beta^*=[1.0,1.0]^\top$ in a fast manner, (2) SGD over OLS with unbiased random label noises would asymptotically converge to a distribution centered at the optimal point, and (3) when $\sigma^2$ increases, the span of the converging distribution becomes larger. In Figure~\ref{fig:SGD-OLS}(e)--(h), we use four settings of $\Sigma_{1,2}$, where we can see (4) no matter how $\Sigma_{1,2}$ is set for OLS problems, the SGD with unbiased random label noises would asymptotically converge to a distribution centered at the optimal point.

Compared the results in~(e) with(f), we can find that, when the trace of $\Sigma_{1,2}$ increases, the span of converging distributions would \lxh{increases}{increase}. Furthermore, (5) the shapes of converging distributions depend on $\Sigma_{1,2}$. In Figure~\ref{fig:SGD-OLS}(g), when we place the principal component of $\Sigma_{1,2}$ onto the vertical axis (i.e., $\Sigma^\mathrm{Ver}=[[10,0]^\top,[0,100]^\top]$), the distribution lays on the vertical axis principally. Figure~\ref{fig:SGD-OLS}(h) demonstrates the opposite layout of the distribution, when we set $\Sigma^\mathrm{Hor}=[[100,0]^\top,[0,10]^\top]$ as $\Sigma_{1,2}$. The scale and shape of the converging distribution backups our theoretical investigation in Eq~\eqref{eq:oup}.

Note that the unbiased random label noises are added to the labels prior to the learning procedure. In this setting, it is the mini-batch sampler of SGD that re-samples the noises and influences the dynamics of SGD through forming the implicit regularizer.

\section{Discussion and Conclusion}
While previous studies primarily focus on the performance degradation caused by label noises or corrupted labels~\cite{jiang2018mentornet,li2020gradient}, we investigate the implicit regularization effects of random label noises, under mini-batch sampling settings of stochastic gradient descent (SGD). 
Specifically, 
we adopt the dynamical systems interpretation of SGD to analyze the learning procedure based on the quadratic loss with unbiased random label noises. We decompose the mini-batch stochastic gradient based on label-noisy losses into three parts in Eq.~(11): (i) $\nabla L^*(\theta)$ -- the true gradient of label-noiseless losses, (ii) $\xi_k^*(\theta)$ -- the stochastic gradient noise caused through mini-batch sampling over the label-noiseless losses, and (iii) $\xi_k^\mathrm{ULN}(\theta)$ -- the noise term influenced by the both random label noises and mini-batch sampling. Our research considers $\xi_k^\mathrm{ULN}(\theta)$ as an implicit regularizer, and finds that effects of such implicit regularizer is to lower the gradient norm of the neural networks $\frac{1}{N}\sum_{i=1}^N\|\nabla_\theta f(x_i,\theta)\|_2^2$ over the learning procedure, where the gradient norm of neural networks here characterizes the variation/stability of the neural network outputs against the random perturbation around the parameters. In summary, the new implicit regularizer $\xi_k^\mathrm{ULN}(\theta)$ helps SGD select a point with higher inference stability for convergence.

We carry out extensive experiments to validate our theoretical investigations. Evaluation based on deep neural network shows that, in self-distillation settings, one can lower the gradient norm of neural networks, improve the inference stability of networks, and obtain better solutions, through iteratively adding noises to the outputs of teacher models. The numerical study with linear regression clearly illustrates the trajectories of SGD-based linear regression with and without unbiased random label noises, the observation coincides the SGD dynamics derived from our theories (i.e., the shape of convergence is controlled by the data, learning rate and batch size). Empirical observations well backup our theoretical findings.

\section*{Declaration}
\begin{itemize}
\item Funding - Haoyi Xiong and Xuhong Li are supported by National Key R\&D Program of China (No. 2021ZD0110303).

\item Conflicts of interest/Competing interests - Not applicable

\item Ethics approval - Not applicable (all based on public available open-source datasets)

\item Consent to participate - Not applicable (all based on public available open-source datasets)

\item Consent for publication - Not applicable (all based on public available open-source datasets)

\item Availability of data and material - All experiments were based on public available open-source datasets. 

\item Code availability - All codes will be released after acceptance.

\item Authors' contributions -
Dr. Haoyi Xiong contributed in the original research ideas, formulated the research problems, proposed part of algorithms, wrote the manuscript and involved in discussions. 
Dr. Xuhong Li contributed in the original research ideas, conducted part of experiments, and wrote part o the manuscript. Mr. Boyang Li helped in part of mathematical proofs. Dongrui Wu and Zhanxing Zhu involved in discussions. Prof. Dejing Dou oversaw the progress of research.
\end{itemize}

\bibliography{main.bib}

\clearpage

\appendix
\section{Proof of Proposition 1}\label{proof:pro1}
\begin{proof}
As the mini-batch $B_k$ are randomly, independently, and uniformly drawn from the full set sample $\mathcal{D}$, thus for $\forall\theta\in\mathbb{R}^d$ and $\forall x_j\in B_k$, there has 
\begin{equation}
  \mathbb{E}_{x_j\in B_k}\left[\nabla L^*_j(\theta)\right]=\frac{1}{N}\sum_{i=1}^N \nabla L_i^*(\theta)\Rightarrow\ \mathbb{E}_{B_k}\ \left[\frac{1}{\vert B_k\vert}\sum_{x_j\in B_k}\nabla L^*_j(\theta) \right]=\frac{1}{N}\sum_{i=1}^N \nabla L^*_i(\theta)\ .
\end{equation}
In this way, we can derive that
\begin{equation}
\begin{aligned}
    \mathbb{E}_{B_k}[\xi^*_k(\theta)] &=  \mathbb{E}_{B_k}\ \sqrt{\eta}\left(\frac{1}{\vert B_k\vert}\sum_{i=1}^k\nabla^*_j(\theta)-\frac{1}{N}\sum_{i=1}^NL^*_i(\theta)\right)\\
    & = \sqrt{\eta}\left(\mathbb{E}_{B_k}\left[\frac{1}{\vert B_k\vert}\sum_{i=1}^k\nabla^*_j(\theta)\right]-\frac{1}{N}\sum_{i=1}^NL^*_i(\theta)\right)
     = \mathbf{0_d}\ .
\end{aligned}
\end{equation}
Further, for any $\theta\in\mathbb{R}^d$ there has 
\begin{equation}
    \begin{aligned}
    \mathrm{Var}[\xi^*_k(\theta)] & =\mathbb{E}_{B_k}\ \left(\xi^*_k(\theta)-\mathbb{E}_{B_k}\xi^*_k(\theta)\right)\left(\xi^*_k(\theta)-\mathbb{E}_{B_k}\xi^*_k(\theta)\right)^\top \\
    & = \mathbb{E}_{B_k} \left[\xi^*_k(\theta)\xi^*_k(\theta)^\top\right]\\
    &=\frac{\eta}{\vert B_k\vert^2} \sum_{x_j\in B_k} \mathrm{Var}[\nabla L^*_j(\theta)]\\
    &= \frac{\eta}{\vert B_k\vert N}\sum_{j=1}^N\left(\nabla L^*_j(\theta)-\frac{1}{N}\sum_{i=1}^NL^*_i(\theta)\right)\left(\nabla L^*_j(\theta)-\frac{1}{N}\sum_{i=1}^NL^*_i(\theta)\right)^\top\\
    & =\frac{\eta}{\vert B_k\vert}\Sigma_N^\mathrm{SGD}(\theta),
    \end{aligned}
\end{equation}
where $\Sigma_N^\mathrm{SGD}(\theta)$ is defined as Eq.~\eqref{eq:covariance}.

Similarly, as the mini-batch $B_k$ are randomly, independently, and uniformly drawn from the full set sample $\mathcal{D}$, there for $\forall\theta\in\mathbb{R}^d$, there has
\begin{equation}
  \mathbb{E}_{x_j\in B_k}\left[\nabla_\theta f(x_j, \theta)\right]=\frac{1}{N}\sum_{i=1}^N \nabla_\theta f(x_i,\theta)\ .
\end{equation}
Thus, there has
\begin{equation}
\begin{aligned}
    \mathbb{E}_{B_k,\varepsilon_i}[\xi_k^\mathrm{ULN}(\theta)]&= -\frac{\sqrt{\eta}}{\vert B_k\vert}\sum_{x_i\in B_k}\left\{\mathbb{E}_{x_i\in B_k}\left[\nabla f(x_i,\theta)]\cdot \mathbb{E}_{\varepsilon_i}[\varepsilon_i\right]\right\}=\mathbf{0_d}\ .
\end{aligned}
\end{equation}
Again, by the definition, there has
\begin{equation}
    \begin{aligned}
    \mathrm{Var}_{B_k,\varepsilon_i}[\xi_k^\mathrm{ULN}(\theta)]&=\mathbb{E}_{B_k,\varepsilon_i}\left[\xi_k^\mathrm{ULN}(\theta)\xi_k^\mathrm{ULN}(\theta)^\top\right]\\
    & \text{Let \lxh{}{us} assume $B_k$ and $\varepsilon_i$ for $1\leq i\leq N$ are independent.}\\
    & = \frac{\eta}{\vert B_k\vert^2}\sum_{x_j\in B_k}\mathbb{E}_{B_k\varepsilon_i}\left[(\varepsilon_i\cdot\nabla_\theta f(x_i,\theta)-\mathbb{E}_{B_k\varepsilon_i}(\varepsilon_i\cdot\nabla_\theta f(x_i,\theta))^2\right]\\
    & \text{As $\mathrm{Var}[\varepsilon_i]=\sigma^2$ for $1\leq i\leq N$, there has}\\
    &=\frac{\eta\sigma^2}{\vert B_k\vert N}\sum_{i=1}^N\nabla_\theta f(x_i,\theta)\nabla_\theta f(x_i,\theta)^\top = \frac{\eta\sigma^2}{\vert B_k\vert}\Sigma_N^\mathrm{ULN}(\theta)\ .
    \end{aligned}
\end{equation}
where $\Sigma_N^\mathrm{ULN}(\theta)$ is defined as Eq.~\eqref{eq:covariance}.
\end{proof}

\section{Proof of Proposition 2}\label{proof:pro2}
\begin{proof}
We show that, as $\eta \rightarrow 0$, the discrete iteration $\bar\theta_k$ of Eq.~\eqref{eq:doublystochdis} in strong norm and on finite--time intervals is close to the solution of the SDE.~\eqref{eq:doublystochcon}.
The main techniques follow~\cite{borkar1999strong}, but~\cite{borkar1999strong} only considered the case when $\Sigma_N^\mathrm{SGD}(\theta)$ and $\Sigma_N^\mathrm{ULN}(\theta)$ are constants.

Let $C_1(\theta)=\sqrt{\frac{1}{B}\Sigma_N^\mathrm{SGD}(\theta)}$, $C_2(\theta)=\sqrt{\frac{1}{B}\Sigma_N^\mathrm{ULN}(\theta)}$ and $L^*(\theta)=\frac{1}{N}\sum_{i=1}^NL_i^*(\theta)$. Let $\widehat{\Theta}(t)$ be the process defined by the integral form of the stochastic differential equation
\begin{equation}\label{eq:HatTheta}
\begin{aligned}
\widehat{\Theta}(t)-\widehat{\Theta}(0) = -\int_0^t \nabla L^*(\widehat{\Theta}_{\lfloor \frac{s}{\eta}\rfloor\eta}) \mathbf{d} s
&+ \sqrt{\eta} \int_0^t C_1(\widehat{\Theta}(\lfloor\frac{s}{\eta}\rfloor\eta)) \mathbf{d} W_1(s)\\
&+ \sqrt{\eta} \int_0^t C_2(\widehat{\Theta}(\lfloor\frac{s}{\eta}\rfloor\eta)) \mathbf{d} W_2(s) \ , \ \widehat{\Theta}(0)=\theta_0 \ .
\end{aligned}
\end{equation}
Here for a real positive number $a>0$ we define $\lfloor a \rfloor=\max\left\{k\in \mathbb{N}_+, k<a\right\}$.
From Eq.\eqref{eq:HatTheta} we see that we have, for $k=0,1,2,...$
\begin{equation}\label{Eq:HatThetaInterpolationSameGLD}
\begin{aligned}
\widehat{\Theta}({(k+1)\eta})-\widehat{\Theta}({k\eta})=&-\eta \nabla L^*(\widehat{\Theta}({k\eta}))-\sqrt{\eta}C_1(\widehat{\Theta}({k\eta}))(W_1((k+1)\eta)-W_1(k\eta)) \\
& - \sqrt{\eta} C_2(\widehat{\Theta}({k\eta}))(W_2((k+1)\eta)-W_2(k\eta)).
\end{aligned}
\end{equation}
Since $\sqrt{\eta}(W_1({(k+1)\eta})-W_1({k\eta}))\sim \mathcal{N}(0, \eta^2 I)$ and $\sqrt{\eta}(W_2({(k+1)\eta})-W_2({k\eta}))\sim \mathcal{N}(0, \eta^2 I)$, 
we could let $\eta z_{k+1}=\sqrt{\eta}(W_1({(k+1)\eta})-W_1({k\eta}))$ and $z'_{k+1}=\sqrt{\eta}(W_2({(k+1)\eta})-W_2({k\eta}))$, where $z_{k+1}$ and $z'_{k+1}$ are the i.i.d. Gaussian sequences in
\eqref{eq:doublystochcon}.
From here, we see that
\begin{equation}\label{Eq:EqualityGLDHatTheta}
\widehat{\Theta}({k\eta})=\bar\theta^\mathrm{ULN}_k \ ,
\end{equation}
where $\bar\theta^\mathrm{ULN}_k$ is the solution to \eqref{eq:doublystochdis}.

We then try to bound $\widehat{\Theta}_t$ in Eq.~\eqref{eq:HatTheta} and $\mathrm{\Theta}^\mathrm{ULB}(t)$ in Eq.~\eqref{eq:doublystochcon}.
Finally we could obtain the error estimation of $\bar\theta^\mathrm{ULN}_k=\widehat{\Theta}({k\eta})$ and $\mathrm{\Theta}^\mathrm{ULN}({k\eta})$ by simply set $t=k\eta$.
Since we assumed that $\nabla L^*_i(\theta)$ and  $\nabla_\theta f(x, \theta)$  are $L$--Lipschitz continuous, we get 
\begin{equation}
\begin{aligned}
\|C_1(\theta_1)-C_1(\theta_2)\|_2&=\sqrt{\frac{1}{bN}\sum_{i=1}^N\|\nabla L^*_i(\theta_1)-\nabla L^*_i(\theta_2)\|_2^2}
\leq L\|\theta_1-\theta_2\|_2\end{aligned}\end{equation}
since the batch size $b\geq 1$.
In the same way, 
\begin{equation}\begin{aligned}
    \|C_2(\theta_1)-C_2(\theta_2)\|_2&=\sqrt{\frac{\sigma^2}{bN}\sum_{i=1}^N\|\nabla_\theta f(x_i,\theta_1)-\nabla_\theta f(x_i,\theta_2)\|_2^2}\leq \sigma L\|\theta_1-\theta_2\|_2\end{aligned}
\end{equation}
since the batch size $B\geq 1$. Thus $C_1(\theta)$ and $C_2$ are both $L$--Lipschitz continuous.
Take a difference between \eqref{eq:HatTheta} and \eqref{eq:doublystochcon} we get
\begin{equation}\label{Eq:DifferenceHatThetaAndTheta}
\begin{aligned}
\widehat{\Theta}(t)-\mathrm{\Theta}^\mathrm{ULN}(t)=&-\int_0^t [\nabla L^*(\widehat\Theta({\lfloor \frac{s}{\eta} \rfloor\eta}))-\nabla L^*(\mathrm{\Theta}^\mathrm{ULN}(s))] \mathbf{d} s\\
&+\sqrt{\eta}\int_0^t [C_1(\widehat{\Theta}({\lfloor \frac{s}{\eta}\rfloor}))
-C_1(\mathrm{\Theta}^\mathrm{ULN}(s))]\mathbf{d} W_1(s)\\
& + \sqrt{\eta}\int_0^t [C_2(\widehat{\Theta}({\lfloor \frac{s}{\eta}\rfloor}))-C_2(\mathrm{\Theta}^\mathrm{ULN}(s))]\mathbf{d} W_2(s).
\end{aligned}
\end{equation}
We can estimate
\begin{equation}\label{Eq:LLipschitzDifferenceGradLossVector}
\begin{aligned}
&\|\nabla L^*(\widehat{\Theta}({\lfloor\frac{s}{\eta}\rfloor\eta}))-\nabla L^*\mathrm{(\Theta}^\mathrm{ULN}(s))\|_2^2 \\
\leq & 2\|\nabla L^*(\widehat{\Theta}({\lfloor\frac{s}{\eta}\rfloor\eta}))-\nabla L^*(\mathrm{\Theta}^\mathrm{ULN}({\lfloor\frac{s}{\eta}\rfloor\eta}))\|_2^2
+ 2\|\nabla L^*(\Theta({\lfloor\frac{s}{\eta}\rfloor\eta}))-\nabla L^*(\mathrm{\Theta}^\mathrm{ULN}(s))\|_2^2
\\
\leq & 2L^2\|\widehat{\Theta}({\lfloor\frac{s}{\eta}\rfloor\eta})-\mathrm{\Theta}^\mathrm{ULN}({\lfloor\frac{s}{\eta}\rfloor\eta})\|_2^2
+2L^2 \|\widehat\Theta(\lfloor\frac{s}{\eta}\rfloor\eta)-\mathrm{\Theta}^\mathrm{ULN}(s)\|_2^2 \ ,
\end{aligned}
\end{equation}
where we used the inequality derived from $L$-Lipschitz. 
%
Similarly, we estimate
\begin{equation}\label{Eq:LLipschitzDifferenceHalfDiffusionMatrix}
\begin{aligned}
& \|C_1(\widehat{\Theta}(\lfloor\frac{s}{\eta}\rfloor\eta))-C_1(\mathrm{\Theta}^\mathrm{ULN}(s))\|_2^2
\\
\leq & 2\|C_1(\widehat{\Theta}(\lfloor\frac{s}{\eta}\rfloor\eta))-C_1(\mathrm{\Theta}^\mathrm{ULN}({\lfloor\frac{s}{\eta}\rfloor\eta}))\|_2^2
+ 2\|C_1(\mathrm{\Theta}^\mathrm{ULN}({\lfloor\frac{s}{\eta}\rfloor\eta}))-\mathrm{\Theta}^\mathrm{ULN}(s))\|_2^2
\\
\leq & 2L^2\|\widehat{\Theta}(\lfloor\frac{s}{\eta}\rfloor\eta)-\mathrm{\Theta}^\mathrm{ULN}({\lfloor\frac{s}{\eta}\rfloor\eta})\|_2^2
+2L^2 \|\mathrm{\Theta}^\mathrm{ULN}({\lfloor\frac{s}{\eta}\rfloor\eta})-\mathrm{\Theta}^\mathrm{ULN}(s))\|_2^2 \ .
\end{aligned}
\end{equation}
In the same way, based on the inequality derived from $L$-Lipschitz, we can also have
\begin{equation}\label{Eq:LLipschitzDifferenceHalfDiffusionMatrix2}
\begin{aligned}
& \|C_2(\widehat{\Theta}(\lfloor\frac{s}{\eta}\rfloor\eta))-C_2(\mathrm{\Theta}^\mathrm{ULN}(s))\|_2^2
\\
\leq & 2\|C_2(\widehat{\Theta}(\lfloor\frac{s}{\eta}\rfloor\eta))-C_2(\mathrm{\Theta}^\mathrm{ULN}({\lfloor\frac{s}{\eta}\rfloor\eta}))\|_2^2
+ 2\|C_2(\mathrm{\Theta}^\mathrm{ULN}({\lfloor\frac{s}{\eta}\rfloor\eta}))-\mathrm{\Theta}^\mathrm{ULN}(s))\|_2^2
\\
\leq & 2\sigma^2L^2\|\widehat{\Theta}(\lfloor\frac{s}{\eta}\rfloor\eta)-\mathrm{\Theta}^\mathrm{ULN}({\lfloor\frac{s}{\eta}\rfloor\eta})\|_2^2
+2\sigma^2L^2 \|\mathrm{\Theta}^\mathrm{ULN}({\lfloor\frac{s}{\eta}\rfloor\eta})-\mathrm{\Theta}^\mathrm{ULN}(s))\|_2^2 \ .
\end{aligned}
\end{equation}

On the other hand, from \eqref{Eq:DifferenceHatThetaAndTheta}, the It\^{o}'s isometry~\cite{oksendal2003stochastic} and Cauchy--Schwarz inequality we have
\begin{equation}\label{Eq:SquareDifferenceHatThetaAndTheta}
\begin{aligned}
 \mathbb{E}\vert\widehat{\Theta}(t)-\mathrm{\Theta}^\mathrm{ULN}(t)\vert^2
\leq & 2 \mathbb{E}\left\|\int_0^t [\nabla L(^*(\Theta({\lfloor \frac{s}{\eta} \rfloor\eta}))-\nabla L^*(\mathrm{\Theta}^\mathrm{ULN}(s))] \mathbf{d}s\right\|_2^2\\
&+2\eta \mathbb{E} \left\|\int_0^t [C_1(\widehat{\Theta}_{\lfloor \frac{s}{\eta}\rfloor})-C_1(\mathrm{\Theta}^\mathrm{ULN}(s))]\mathbf{d}W_1(s)\right\|^2\\
&+2\eta \mathbb{E} \left\|\int_0^t [C_2(\widehat{\Theta}_{\lfloor \frac{s}{\eta}\rfloor})-C_2(\mathrm{\Theta}^\mathrm{ULN}(s))]\mathbf{d}W_1(s)\right\|^2\\
\leq &2\int_0^t  \mathbb{E}\left\|\nabla L(^*(\Theta({\lfloor \frac{s}{\eta} \rfloor\eta}))-\nabla L^*(\mathrm{\Theta}^\mathrm{ULN}(s))\right\|_2^2\mathbf{d}s\\
&+2\eta \int_0^t  \mathbb{E}\left\|C_1(\widehat{\Theta}_{\lfloor \frac{s}{\eta}\rfloor})-C_1(\mathrm{\Theta}^\mathrm{ULN}(s))\right\|^2\mathbf{d}s\\
&+2\eta \int_0^t  \mathbb{E}\left\|C_2(\widehat{\Theta}_{\lfloor \frac{s}{\eta}\rfloor})-C_2(\mathrm{\Theta}^\mathrm{ULN}(s))\right\|^2\mathbf{d}s \ .
\end{aligned}
\end{equation}
Combining Eqs.~\eqref{Eq:LLipschitzDifferenceGradLossVector}, \eqref{Eq:LLipschitzDifferenceHalfDiffusionMatrix}, \eqref{Eq:LLipschitzDifferenceHalfDiffusionMatrix2} and
\eqref{Eq:SquareDifferenceHatThetaAndTheta} we obtain that
\begin{equation}\label{Eq:SquareDifferenceHatThetaAndThetaGronwall-1}
\begin{aligned}
&  \mathbb{E}\|\widehat{\Theta}(t)-\mathrm{\Theta}^\mathrm{ULN}(t)\|_2^2
\\
 &\leq
2\int_0^t \left(2L^2 \mathbb{E} \|\widehat{\Theta}(\lfloor\frac{s}{\eta}\rfloor\eta)-\mathrm{\Theta}^\mathrm{ULN}({\lfloor\frac{s}{\eta}\rfloor\eta})\|_2^2+2L^2 \mathbb{E}\|\Theta_{\lfloor \frac{s}{\eta} \rfloor\eta}-\mathrm{\Theta}^\mathrm{ULN}(s))\|_2^2\right)\mathbf{d}s
\\
&+2\eta \int_0^t \left(2L^2 \mathbb{E}\|\widehat{\Theta}(\lfloor\frac{s}{\eta}\rfloor\eta)-\mathrm{\Theta}^\mathrm{ULN}({\lfloor\frac{s}{\eta}\rfloor\eta})\|_2^2
+2L^2  \mathbb{E} \|\mathrm{\Theta}^\mathrm{ULN}({\lfloor\frac{s}{\eta}\rfloor\eta})-\mathrm{\Theta}^\mathrm{ULN}(s))\|_2^2\right)\mathbf{d}s \\
& +2\eta \int_0^t \left(2\sigma^2L^2 \mathbb{E}\|\widehat{\Theta}(\lfloor\frac{s}{\eta}\rfloor\eta)-\mathrm{\Theta}^\mathrm{ULN}({\lfloor\frac{s}{\eta}\rfloor\eta})\|_2^2
+2\sigma^2L^2  \mathbb{E} \|\mathrm{\Theta}^\mathrm{ULN}({\lfloor\frac{s}{\eta}\rfloor\eta})-\mathrm{\Theta}^\mathrm{ULN}(s))\|_2^2\right)\mathbf{d}s \ .
\\
=&4(1+\eta+\eta\sigma^2) L^2\cdot \\
&\left(\int_0^t  \mathbb{E}\|\widehat{\Theta}(\lfloor\frac{s}{\eta}\rfloor\eta)-\mathrm{\Theta}^\mathrm{ULN}({\lfloor\frac{s}{\eta}\rfloor\eta})\|_2^2\mathbf{d}s
+\int_0^t  \mathbb{E} \|\mathrm{\Theta}^\mathrm{ULN}({\lfloor\frac{s}{\eta}\rfloor\eta})-\mathrm{\Theta}^\mathrm{ULN}(s))\|_2^2 \mathbf{d}s\right) \ .
\end{aligned}
\end{equation}

Since we assumed that there is an $M>0$ such that $\max_{i=1,2...,N}(\vert\nabla L^*_i(\theta)\vert)\leq M$, we conclude that $\vert\nabla L^*(\theta)\vert\leq \dfrac{1}{N}\sum_{i=1}^N \vert\nabla L^*_i(\theta)\vert\leq M$ and 
\begin{equation}
    \|C_1(\theta)\|_2\leq \dfrac{1}{\sqrt{b}}\sqrt{\frac{1}{N}\sum_{i=1}^N \|\nabla L^*_i(\theta)\|_2^2}\leq M,
\end{equation}
\begin{equation}
    \|C_2(\theta)\|_2\leq \dfrac{1}{\sqrt{b}}\sqrt{\frac{\sigma^2}{N}\sum_{i=1}^N \|\nabla_\theta f(x_i,\theta)\|_2^2}\leq \sigma M
\end{equation}
since $B\geq 1$. 
By Eq.~\eqref{eq:doublystochcon}, the It\^{o}'s isometry~\cite{oksendal2003stochastic}, the Cauchy-Schwarz inequality
and $0\leq s-\lfloor\frac{s}{\eta}\rfloor\eta \leq \eta$ we know that

\begin{equation}\label{Eq:SquareDifferenceThetaDiscretization}
\begin{aligned}
&   \mathbb{E} \|\mathrm{\Theta}^\mathrm{ULN}({\lfloor\frac{s}{\eta}\rfloor\eta})-\mathrm{\Theta}^\mathrm{ULN}(s))\|_2^2
\\
= & \displaystyle{ \mathbb{E} \left\|-\int_{\lfloor\frac{s}{\eta}\rfloor\eta}^s \nabla L^*(\Theta_u) \mathbf{d}u
+\sqrt{\eta}\int_{\lfloor\frac{s}{\eta}\rfloor\eta}^s C_1(\Theta_u)\mathbf{d}W_1(u) 
+\sqrt{\eta}\int_{\lfloor\frac{s}{\eta}\rfloor\eta}^s C_2(\Theta_u)\mathbf{d}W_2(u)\right\|_2^2}
\\
\leq & \displaystyle{2 \mathbb{E} \left\|\int_{\lfloor\frac{s}{\eta}\rfloor\eta}^s \nabla L^*(\Theta_u) \mathbf{d}u\right\|_2^2
+2\eta \mathbb{E}\left\|\int_{\lfloor\frac{s}{\eta}\rfloor\eta}^s C_1(\Theta_u)\mathbf{d}W_1(u) \right\|_2^2
+2\eta \mathbb{E}\left\|\int_{\lfloor\frac{s}{\eta}\rfloor\eta}^s C_2(\Theta_u)\mathbf{d}W_2(u) \right\|_2^2}
\\
\leq & 2 \mathbb{E} \left(\int_{\lfloor\frac{s}{\eta}\rfloor\eta}^s \left\|\nabla L^*(\Theta_u)\right\|\mathbf{d}u\right)^2
+2\eta\int_{\lfloor\frac{s}{\eta}\rfloor\eta}^s  \mathbb{E}\|C_1(\Theta_u)\|_2^2 \mathbf{d}u
+2\eta\int_{\lfloor\frac{s}{\eta}\rfloor\eta}^s  \mathbb{E}\|C_2(\Theta_u)\|_2^2 \mathbf{d}u
\\
\leq & \displaystyle{2\eta \int_{\lfloor\frac{s}{\eta}\rfloor\eta}^s  \mathbb{E}\vert\nabla L^*(\Theta_u)\vert^2 \mathbf{d}u
+2\eta \int_{\lfloor\frac{s}{\eta}\rfloor\eta}^s  \mathbb{E}\|C_1(\Theta_u)\|_2^2 \mathbf{d}u
+2\eta \int_{\lfloor\frac{s}{\eta}\rfloor\eta}^s  \mathbb{E}\|C_2(\Theta_u)\|_2^2 \mathbf{d}u}
\\
\leq & 2\eta^2 M^2
+2\eta^2 M^2+2\eta^2\sigma^2 M^2=(4+2\sigma^2)\eta^2 M^2 \ .
\end{aligned}
\end{equation}

Combining \eqref{Eq:SquareDifferenceThetaDiscretization} and \eqref{Eq:SquareDifferenceHatThetaAndThetaGronwall-1}
we obtain
\begin{equation}
\begin{aligned}
 &\mathbb{E}\|\widehat{\Theta}(t)-\mathrm{\Theta}^\mathrm{ULN}(t)\|_2^2\\
\leq&
4(1+\eta+\eta\sigma^2)L^2 \cdot \left(\int_0^t  \mathbb{E}\|\widehat{\Theta}(\lfloor\frac{s}{\eta}\rfloor\eta)-\mathrm{\Theta}^\mathrm{ULN}({\lfloor\frac{s}{\eta}\rfloor\eta})\|_2^2\mathbf{d}s
+(4+2\sigma^2)\eta^2 M^2 t\right) \ .
\end{aligned}
\end{equation}

Set $T>0$ and $m(t)=\max\limits_{0\leq s\leq t} \mathbb{E} \vert\widehat{\Theta}_s-\mathrm{\Theta}^\mathrm{ULN}(s))\vert^2$, noticing that
$m(\lfloor\frac{s}{\eta}\rfloor \eta)\leq m(s)$ (as $\lfloor\frac{s}{\eta}\rfloor\eta \leq s$), then the above gives for any $0\leq t\leq T$,
\begin{equation}\label{Eq:SquareDifferenceHatThetaAndThetaGronwall-2}
m(t) \leq
4(1+\eta+\eta\sigma^2)L^2 \cdot \left(\int_0^t m(s)\mathbf{d}s
+(4+2\sigma^2)\eta^2 M^2 T\right) \ .
\end{equation}

By Gronwall's inequality we obtain that for $0\leq t \leq T$,
\begin{equation}
m(t) \leq
4(1+\eta+\eta\sigma^2)(4+2\sigma^2) L^2\eta^2 M^2 T e^{4(1+\eta+\eta\sigma^2) L^2 t}.
\end{equation}
Suppose $0<\eta<1$, then there is a constant $C$ which is independent on $\eta$ s.t.
\begin{equation}\label{Eq:SquareDifferenceHatThetaAndThetaGronwall-Final}
     \mathbb{E}\|\widehat{\Theta}(t)-\mathrm{\Theta}^\mathrm{ULN}(t)\|_2^2 \le m(t) \leq C \eta^2.
\end{equation}
Set $t=k\eta$ in \eqref{Eq:SquareDifferenceHatThetaAndThetaGronwall-Final} and make use of \eqref{Eq:EqualityGLDHatTheta},
we finish the proof.
\end{proof}

\section{Proof of Proposition 3}\label{proof:pro3}
\begin{proof}
To obtain Eq~\eqref{eq:strength}, we can use the simple vector-matrix-vector products transform that, for the random vector $v$ and symmetric matrix $A$ there has $\mathbb{E}_v [v^\top Av]=\mathrm{trace}(A\mathbb{E}[vv^\top])$, such that
\begin{equation}
\begin{aligned}
\mathbb{E}_{z_k}\|\xi_k^{\mathrm{ULN}}(\theta)\|_2^2
&=\mathbb{E}_{z_k}\left[\xi_k^{\mathrm{ULN}}(\theta)^\top \xi_k^{\mathrm{ULN}}(\theta)\right]\\
&\approx{\frac{\eta\sigma^2}{b}}\mathbb{E}_{z_k}\left[z_k^\top\left(\frac{1}{N}\sum_{i=1}^N\nabla_\theta f(x_i,\theta)\nabla_\theta f(x_i,\theta)^\top\right)z_k\right]\\
&={\frac{\eta\sigma^2}{b}}\mathrm{trace}\left(\left(\frac{1}{N}\sum_{i=1}^N\nabla_\theta f(x_i,\theta)\nabla_\theta f(x_i,\theta)^\top\right)\mathbb{E}_{z_k}[z_kz_k^\top]\right)\\
& \text{ as } \mathbb{E}_{z_k}[z_kz_k^\top]=\mathbf{I}_d\ for\ z_k\sim\mathcal{N}(\mathbf{0}_d,\mathbf{I}_d) \\
&=\frac{\eta\sigma^2}{b}\mathrm{trace}\left(\frac{1}{N}\sum_{i=1}^N\nabla_\theta f(x_i,\theta)\nabla_\theta f(x_i,\theta)^\top\right)\\
&=\frac{\eta\sigma^2}{bN}\sum_{i=1}^N\left\|\nabla_\theta f(x_i,\theta)\right\|_2^2
.
\end{aligned}
\end{equation}
\end{proof}


\end{document}